\documentclass{article}

\usepackage{PRIMEarxiv}

\usepackage[utf8]{inputenc} % allow utf-8 input
\usepackage[T1]{fontenc}    % use 8-bit T1 fonts
\usepackage{hyperref}       % hyperlinks
\usepackage{url}            % simple URL typesetting
\usepackage{booktabs}       % professional-quality tables
\usepackage{amsfonts}       % blackboard math symbols
\usepackage{nicefrac}       % compact symbols for 1/2, etc.
\usepackage{microtype}      % microtypography
\usepackage{lipsum}
\usepackage{fancyhdr}       % header
\usepackage{graphicx}       % graphics
\graphicspath{{media/}}     % organize your images and other figures under media/ folder

\usepackage{xcolor}
\usepackage{colortbl}
\usepackage{siunitx}
\usepackage{multirow}
\usepackage{rotating}
\usepackage{adjustbox}
\usepackage{amsmath}
\usepackage{subcaption}

\definecolor{lightgray}{gray}{0.9}
\definecolor{bestresult}{rgb}{0.9,1.0,0.9}

\title{EAGLE: Efficient Alignment of Generalized Latent Embeddings for Multimodal Survival Prediction with Interpretable Attribution Analysis}

\author{
  Aakash Tripathi$^{1,3,*}$ \\
  Dept. of Machine Learning \\
  Moffitt Cancer Center \\
  Tampa, FL \\
  \texttt{aakash.tripathi@moffitt.org} \\
  \And
  Asim Waqas$^{1,2,*}$ \\
  Dept. of Machine Learning \\
  Moffitt Cancer Center \\
  Tampa, FL \\
  \texttt{asim.waqas@moffitt.org} \\
  \And
  Matthew B. Schabath$^{2}$ \\
  Dept. of Cancer Epidemiology \\
  Moffitt Cancer Center \\
  Tampa, FL \\
  \texttt{matthew.schabath@moffitt.org} \\
  \AND
  Yasin Yilmaz$^{3}$ \\
  Dept. of Electrical Engineering \\
  University of South Florida \\
  Tampa, FL \\
  \texttt{yasiny@usf.edu} \\
  \And
  Ghulam Rasool$^{1,3}$ \\
  Dept. of Machine Learning \\
  Moffitt Cancer Center \\
  Tampa, FL \\
  \texttt{ghulam.rasool@moffitt.org} \\
}

\begin{document}
\maketitle

\begin{center}
\vspace{-0.5cm}
\footnotesize
$^*$Equal contribution \\[0.3ex]
$^1$Department of Machine Learning, Moffitt Cancer Center \& Research Institute \\
$^2$Department of Cancer Epidemiology, Moffitt Cancer Center \& Research Institute \\
$^3$Department of Electrical Engineering, University of South Florida
\vspace{0.3cm}
\end{center}

\begin{abstract}
Accurate cancer survival prediction requires integration of diverse data modalities that reflect the complex interplay between imaging, clinical parameters, and textual reports. However, existing multimodal approaches suffer from simplistic fusion strategies, massive computational requirements, and lack of interpretability—critical barriers to clinical adoption. We present EAGLE (Efficient Alignment of Generalized Latent Embeddings), a novel deep learning framework that addresses these limitations through attention-based multimodal fusion with comprehensive attribution analysis. EAGLE introduces four key innovations: (1) dynamic cross-modal attention mechanisms that learn hierarchical relationships between modalities, (2) massive dimensionality reduction (99.96\%) while maintaining predictive performance, (3) three complementary attribution methods providing patient-level interpretability, and (4) a unified pipeline enabling seamless adaptation across cancer types. We evaluated EAGLE on 911 patients across three distinct malignancies: glioblastoma (GBM, n=160), intraductal papillary mucinous neoplasms (IPMN, n=171), and non-small cell lung cancer (NSCLC, n=580). EAGLE achieved concordance indices of 0.637±0.087 (GBM), 0.679±0.029 (IPMN), and 0.598±0.021 (NSCLC), demonstrating robust performance across diverse cancer types. Attribution analysis revealed disease-specific patterns: text reports dominated GBM predictions (43.7\%), balanced contributions characterized IPMN (31-35\% per modality), while imaging drove NSCLC risk assessment (49.0\%). Patient-level analysis showed high-risk individuals relied more heavily on adverse imaging features, while low-risk patients demonstrated balanced modality contributions. Risk stratification identified clinically meaningful groups with 4-fold (GBM) to 5-fold (NSCLC) differences in median survival, directly informing treatment intensity decisions. By combining state-of-the-art performance with clinical interpretability, EAGLE bridges the gap between advanced AI capabilities and practical healthcare deployment, offering a scalable solution for multimodal survival prediction that enhances both prognostic accuracy and physician trust in automated predictions.
\end{abstract}

% keywords can be removed
\keywords{Multimodal \and Oncology \and Embeddings \and Overall Survival \and Framework}

\section{Introduction}

The landscape of precision medicine has undergone a fundamental transformation as healthcare systems increasingly recognize that accurate survival prediction requires comprehensive integration of diverse data modalities that clinicians routinely encounter in practice. Traditional prognostic models, often limited to single data sources, fail to capture the complex interplay between anatomical structures visualized through medical imaging, quantitative clinical parameters recorded in electronic health records, and nuanced observations documented in radiology and pathology reports. This gap between clinical reality and computational modeling has motivated the development of multimodal survival prediction frameworks that mirror the integrative decision-making process of experienced oncologists.

The imperative for multimodal survival prediction emerges from profound clinical challenges across multiple cancer types, where single-modality approaches consistently underperform in capturing disease complexity. Glioblastoma multiforme (GBM), the most aggressive primary brain tumor, exemplifies this challenge with median survival times of 12-15 months and fewer than 10\% of patients surviving beyond three years despite aggressive treatment \cite{lu2016molecular}. Current prognostic models for GBM show variable performance, with concordance indices ranging from 0.6-0.74, highlighting the need for improved discriminative ability in personalized treatment planning \cite{hajianfar2023time}. Similarly, non-small cell lung cancer (NSCLC) presents substantial heterogeneity, with 52\% of patients presenting with distant metastases at diagnosis and five-year survival rates varying dramatically from 10\% for metastatic disease to 39.4\% for localized tumors \cite{li2025survival}. Traditional staging systems prove inadequate for capturing this variability, necessitating more sophisticated approaches to guide critical treatment decisions \cite{cam44946}. The clinical dilemma extends to intraductal papillary mucinous neoplasms (IPMN), where only a minority progress to invasive carcinoma, yet 58.2\% of patients experience disease progression during surveillance with a 72.1\% cumulative risk for surgery at 10 years \cite{del2017survival}. This prognostic uncertainty directly impacts patient care, as invasive IPMNs demonstrate five-year survival rates of 24-40\% compared to over 90\% for non-invasive lesions \cite{kang2020risk}. These stark differences in outcomes underscore the critical need for more accurate prediction models that can distinguish between indolent and aggressive disease phenotypes.

Clinical practice inherently operates in a multimodal paradigm, with physicians integrating medical imaging (CT, MRI, PET), clinical parameters (laboratory values, vital signs, demographics), molecular information (genomics, transcriptomics), and textual documentation (clinical notes, pathology reports) to formulate prognoses \cite{kumar2024multimodality}. Evidence suggests that multimodal approaches can outperform single-modality models, with studies reporting varying degrees of improvement depending on the specific application and cancer type \cite{kline2022multimodal}. The MultiSurv framework achieved a C-index of 0.822 across 33 cancer types compared to 0.784 for single-modality approaches, suggesting the potential value of comprehensive data integration for improving prognostic accuracy \cite{vale2021long}. Beyond technical performance metrics, multimodal survival prediction addresses critical clinical needs including personalized treatment selection, resource optimization, and quality of life considerations. Studies indicate that physicians overestimate survival in 27\% of advanced cancer cases by four or more weeks, highlighting the potential for objective, data-driven models to improve prognostic accuracy and support difficult clinical conversations \cite{cui2014predicting}. Healthcare systems benefit through optimized resource allocation, reduced unnecessary treatments in poor-prognosis patients, and standardized prognostic assessments across institutions \cite{banerji2025clinicians}.

Despite growing recognition of multimodal integration's importance, existing survival prediction models exhibit fundamental limitations that restrict their clinical utility. Single-modality approaches provide incomplete patient representations, missing crucial complementary information that influences prognosis. Imaging-only models cannot account for patient-specific factors such as comorbidities or performance status, while clinical-only models miss morphological features and spatial relationships within tumors \cite{li2024review}. Genomics-only approaches ignore clinical context and the tumor microenvironment, potentially resulting in reduced predictive performance compared to multimodal alternatives \cite{10.5555/3692070.3693942}. Current multimodal fusion strategies often rely on approaches that may not fully capture complex inter-modal relationships. Many existing methods employ feature concatenation, which assumes equal importance across modalities and may not capture their hierarchical relationships \cite{li2024review}. Late fusion approaches combine predictions from independently trained models, potentially missing opportunities to learn cross-modal interactions during training \cite{schouten2025navigating}. These fusion strategies may lack adaptability to varying data quality or missing modalities, leading to degraded performance when complete multimodal data is unavailable—a common scenario in clinical practice.

Technical constraints further limit existing approaches' scalability and generalizability. Many frameworks are designed for specific cancer types or imaging protocols, requiring substantial re-engineering for new applications \cite{meng2024adaptive}. The absence of unified training and evaluation pipelines across different cancer types results in inconsistent performance metrics and difficulty comparing approaches. Missing data handling remains rudimentary, with most models requiring complete multimodal inputs or suffering significant performance degradation with incomplete data \cite{liu2023attention}. Additionally, the black-box nature of deep multimodal models poses interpretability challenges that hinder clinical adoption, as physicians require understanding of which features drive predictions to trust and act upon model outputs \cite{amann2020explainability}.

The evolution of survival prediction has progressed through distinct waves of methodological advancement, each addressing specific limitations of previous approaches. Imaging-based survival analysis has transitioned from traditional radiomics to sophisticated deep learning architectures. Early approaches extracted hand-crafted features—including first-order statistics, texture measures, and shape descriptors—combined with Cox proportional hazards models \cite{kourou2015machine}. Modern frameworks employ 3D convolutional neural networks (CNNs), with architectures like M²Net combining multi-modal multi-channel networks for brain tumor survival prediction \cite{gomaa2024comprehensive}. The DRAG model utilizes 3D U-Net architectures for feature extraction, while recent implementations incorporate vision transformers (ViT) with self-attention mechanisms to capture long-range dependencies in medical images \cite{meng2024adaptive}. These approaches show promise in improving upon traditional radiomics, though opportunities remain for further advancement through multimodal integration.

Clinical data survival models have similarly evolved from traditional Cox proportional hazards to sophisticated deep learning frameworks. DeepSurv extends Cox models with neural networks to capture nonlinear relationships, achieving C-indices of 0.769-0.798 across various datasets \cite{katzman2018deepsurv}. DeepHit abandons proportional hazards assumptions entirely, employing discrete-time modeling with cause-specific subnetworks \cite{lee2018deephit}. Recent innovations include Dynamic-DeepHit for longitudinal data, SurvTRACE leveraging transformer architectures, and Deep Survival Machines for non-parametric modeling \cite{wiegrebe2024deep}. These methods increasingly incorporate attention mechanisms for clinical variable importance weighting and sophisticated missing data strategies including imputation, masking, and learnable embeddings.

Multimodal fusion techniques represent an active area of research in survival prediction, with architectures evolving from simple concatenation to sophisticated attention-based mechanisms. Early fusion approaches combine raw inputs before feature extraction but may lose modality-specific characteristics \cite{waqas2024multimodal}. Late fusion methods ensemble predictions from independently trained models but may miss critical cross-modal interactions during training. The emergence of transformer-based architectures has advanced multimodal medical AI, with frameworks like TransMed combining CNNs and transformers, IRENE employing bidirectional multimodal attention blocks, and 3MT utilizing cascaded modality transformers with cross-attention \cite{simon2024future}. Graph neural network approaches, including PathomicFusion for morphological-genomic integration, provide alternative paradigms for modeling complex relationships between modalities \cite{chen2020pathomic}. Advanced methods increasingly employ contrastive learning for robust representations, self-supervised pretraining on unlabeled data, and foundation model adaptation for medical tasks.

The EAGLE framework addresses key challenges in existing multimodal survival prediction through four main contributions that aim to balance technical innovation with clinical applicability. 

(1) EAGLE introduces an attention-based fusion architecture that dynamically learns cross-modal interactions through bidirectional attention mechanisms, enabling the model to adaptively weight modality contributions based on their relevance to specific patients and cancer types. Unlike traditional concatenation or late fusion approaches, this architecture aims to capture relationships between imaging features, clinical parameters, and textual information through learned attention weights that adjust to data quality and availability.

(2) EAGLE implements interpretable attribution analysis at both instance and modality levels, addressing the need for explainability in clinical decision support. The framework integrates multiple attribution techniques—including simple magnitude-based attribution, gradient-based methods, and integrated gradients—providing clinicians with explanations of prediction rationale. This interpretability extends beyond simple feature importance to reveal modality contributions that influence survival predictions, enabling physicians to better understand model reasoning.

(3) EAGLE establishes a unified training and evaluation pipeline that standardizes multimodal survival prediction across multiple cancer types (GBM, IPMN, NSCLC), aiming for generalizability beyond disease-specific frameworks. This unified approach employs consistent preprocessing protocols, including standardization for imaging data, robust imputation strategies for clinical variables, and domain-specific embeddings for textual reports using medical language models. The framework's modular architecture enables adaptation to new cancer types while maintaining consistent evaluation protocols.

(4) EAGLE achieves substantial dimensionality reduction (over 99\%) from the original multimodal embeddings while maintaining competitive predictive performance with traditional survival models. The framework's training strategy incorporates standard techniques to handle varying data availability, aiming to maintain robust performance when dealing with real-world clinical data. Combined with established training procedures for survival analysis, EAGLE provides a practical framework for multimodal survival prediction that balances model complexity with computational efficiency.

\section{Methodology}

EAGLE is a multimodal deep learning framework designed to integrate imaging, clinical, and textual data for cancer patient survival prediction. The framework employs attention-based fusion mechanisms to learn cross-modal interactions while maintaining interpretability through attribution analysis. Our approach addresses three key challenges: (1) effective integration of heterogeneous data modalities, (2) substantial dimensionality reduction while maintaining model performance, and (3) clinical interpretability of model predictions.

\begin{figure}
    \centering
    \includegraphics[width=1\linewidth]{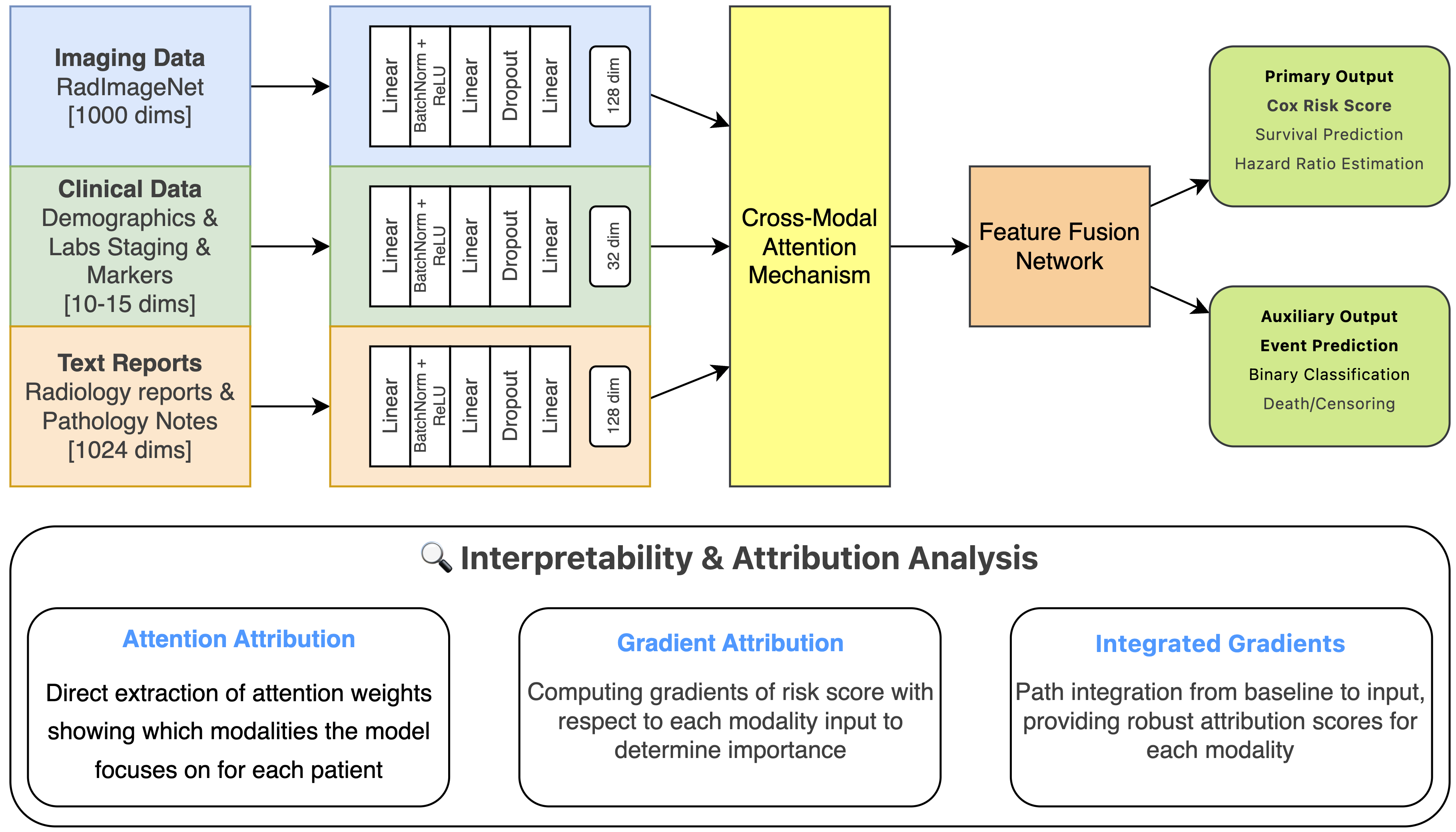}
    \caption{EAGLE architecture for multimodal survival prediction. The framework processes three heterogeneous data modalities: (1) imaging data using RadImageNet embeddings (1000-2048 dimensions), (2) clinical data including demographics, laboratory values, staging, and molecular markers (10-36 dimensions), and (3) text reports from radiology and pathology (768-2304 dimensions from GatorTron embeddings). Each modality passes through dedicated encoder networks consisting of linear transformations, batch normalization, ReLU activations, and dropout layers, reducing dimensionality to 128 (imaging/text) or 32 (clinical) dimensions. The encoded representations undergo cross-modal attention fusion to capture inter-modality relationships before final feature fusion. The network produces two outputs: a primary Cox risk score for survival prediction and an auxiliary binary classification for event prediction. The framework incorporates three complementary attribution methods—simple magnitude-based attribution, gradient-based attribution, and integrated gradients—to provide interpretable explanations of modality contributions at both patient and population levels. This architecture achieves approximately 97-98\% dimensionality reduction from the pre-extracted embeddings while aiming to maintain predictive performance across cancer types.}
    \label{fig:eagle}
\end{figure}

\subsection{Datasets and Preprocessing}

We evaluated EAGLE on three distinct cancer cohorts representing diverse anatomical sites and clinical characteristics:

\subsubsection{Glioblastoma (GBM)}

The GBM dataset comprises 160 patients with histologically confirmed glioblastoma multiforme. All patients had complete multimodal data including:
\begin{itemize}
    \item \textbf{Imaging}: MRI sequences (T1-weighted with/without contrast, T2-weighted, FLAIR) with 155 slices per patient, processed into 1000-dimensional RadImageNet embeddings
    \item \textbf{Clinical features}: Age at diagnosis (mean: 62.6 years, range: 21-91), gender (60\% male), race, ethnicity, tumor size (mean: 10.8 cm), and treatment details
    \item \textbf{Text reports}: Radiology reports (100\% availability), pathology reports (96.2\% availability), and treatment summaries
    \item \textbf{Outcomes}: Median survival of 13.0 months (395.7 days) with a 95.6\% event rate, reflecting the aggressive nature of GBM
\end{itemize}

\subsubsection{Intraductal Papillary Mucinous Neoplasm (IPMN)}

The IPMN dataset includes 171 patients with pancreatic cystic lesions. The cohort characteristics include:
\begin{itemize}
    \item \textbf{Imaging}: Triple-phase CT scans (arterial, venous, non-contrast) available for 87.1\% of patients, with 190 slices per patient processed into 1000-dimensional embeddings
    \item \textbf{Clinical features}: BMI (mean: 27.5), smoking history (42.7\% ever smokers), serum CA 19-9 levels (median: 15.8 U/mL, available for 91.8\%), histology, and grade
    \item \textbf{Text reports}: Radiology reports (78.9\% availability) and pathology reports (98.2\% availability)
    \item \textbf{Outcomes}: Median survival of 6.53 years (2,384 days) with a 48.5\% event rate, reflecting the heterogeneous nature of IPMNs ranging from benign to malignant
\end{itemize}

\subsubsection{Non-Small Cell Lung Cancer (NSCLC)}

The NSCLC dataset represents our largest cohort with 580 patients:
\begin{itemize}
    \item \textbf{Imaging}: Dual CT imaging with contrast-enhanced CT (74.0\% availability) and non-contrast CT (37.2\% availability), each producing 1024-dimensional embeddings
    \item \textbf{Clinical features}: Age at diagnosis (mean: 69.6 years), race, ethnicity, smoking history (81.9\% ever smokers), TNM staging (clinical and pathological), and histology (38.8\% adenocarcinoma)
    \item \textbf{Text reports}: Clinical embeddings incorporating radiology and clinical documentation
    \item \textbf{Outcomes}: Median survival of 35.0 months (1,065 days) with a 51.7\% event rate
\end{itemize}

\subsubsection{Data Preprocessing}

Each modality underwent specific preprocessing:

\textbf{Imaging Data}: We utilized pre-extracted embeddings from medical images using domain-specific models (RadImageNet for general imaging features). For datasets with multiple imaging sequences (e.g., contrast and non-contrast CT in NSCLC), embeddings were concatenated to preserve complementary information. This approach reduces computational requirements while leveraging pre-trained medical imaging representations.

\textbf{Clinical Features}: Numerical features were standardized using median imputation for missing values followed by z-score normalization. Categorical features were encoded using label encoding with an "Unknown" category for missing values. Missing data rates varied: 9.4\% for tumor size in GBM, 8.2\% for CA 19-9 in IPMN, and approximately 47\% for pathological staging in NSCLC.

\textbf{Text Reports}: Natural language reports (radiology, pathology, treatment notes) were processed using GatorTron, a clinical language model, to generate embeddings. For GBM and IPMN, multiple report types were concatenated, resulting in 2304 and 1536 dimensions respectively. Additionally, dataset-specific feature extractors identified clinically relevant binary features (e.g., presence of MGMT methylation in GBM reports, main duct involvement in IPMN).

\subsection{Model Architecture}

EAGLE consists of four main components: modality-specific encoders, dimension harmonization layers, attention-based fusion, and survival prediction heads (Figure \ref{fig:eagle}).

\subsubsection{Modality-Specific Encoders}

Each modality is processed through dedicated encoder networks that transform pre-extracted embeddings into learned representations. The architecture varies by dataset to accommodate different data characteristics:

\textbf{Imaging Encoder}: 
\begin{itemize}
    \item Default architecture: [512, 256, 128] with batch normalization, ReLU activation, and dropout (p=0.3)
    \item IPMN-specific: [256, 128] to accommodate smaller cohort size
    \item Input dimensions: 1000 (GBM/IPMN) or 2048 (NSCLC with dual CT)
\end{itemize}

\textbf{Text Encoder}:
\begin{itemize}
    \item Default architecture: [512, 256, 128] for GBM
    \item NSCLC-specific: [256, 128] for single clinical embedding
    \item IPMN-specific: [256, 128] for concatenated reports
    \item Input dimensions vary: 2304 (GBM), 1536 (IPMN), 1024 (NSCLC)
\end{itemize}

\textbf{Clinical Encoder}: A smaller network [64, 32] or [128, 64, 32] (NSCLC) designed for lower-dimensional clinical features, following the same layer structure as other encoders.

\subsubsection{Attention-Based Fusion}

To capture cross-modal interactions, we employ multi-head attention mechanisms (Figure \ref{fig:eagle}). After encoding, all modality representations are projected to a common dimension through linear transformation layers. The attention fusion process operates as follows:

\begin{enumerate}
    \item Modality embeddings are arranged as sequences for attention computation
    \item Cross-modal attention is applied between imaging-text and imaging-clinical pairs
    \item The attention mechanism uses 8 heads with dropout (p=0.1) for regularization
    \item Attended features are pooled and concatenated for final fusion
\end{enumerate}

The attention weights are preserved during forward passes to enable attribution analysis.

\subsubsection{Survival Prediction}

The fused multimodal representation passes through a final fusion network before the prediction heads:

\textbf{Primary Task}: A Cox regression head that outputs a single risk score for survival analysis

\textbf{Auxiliary Task}: A binary classification head predicting event occurrence, trained jointly to improve feature learning

The fusion network architecture is [256, 128, 64] by default but may be adjusted based on dataset characteristics.

\subsection{Loss Function}

The model is trained using the Cox partial likelihood loss, which handles censored survival data appropriately. For a batch of patients sorted by survival time in descending order, the loss is computed as:

$$\mathcal{L}_{cox} = -\frac{1}{N_{events}} \sum_{i: \delta_i = 1} \left[ \beta^T x_i - \log\left(\sum_{j \in R_i} \exp(\beta^T x_j)\right) \right]$$

where $\delta_i$ is the event indicator, $R_i$ is the risk set at time $t_i$, and $\beta^T x_i$ represents the model's risk score for patient $i$.

The total loss combines the Cox loss with the auxiliary event prediction loss:

$$\mathcal{L}_{total} = \mathcal{L}_{cox} + \lambda \cdot \mathcal{L}_{event}$$

where $\lambda = 0.1$ weights the auxiliary task contribution.

\subsection{Training Procedure}

Models were trained using 5-fold stratified cross-validation, with stratification based on event status to maintain class balance across folds. Training configurations were adapted for each dataset:

\begin{itemize}
    \item \textbf{Optimizer}: AdamW with dataset-specific learning rates (1e-4 for GBM/IPMN, 5e-5 for NSCLC) and weight decay 0.01
    \item \textbf{Scheduler}: ReduceLROnPlateau monitoring validation C-index
    \item \textbf{Early Stopping}: Patience of 15 epochs based on validation performance
    \item \textbf{Gradient Clipping}: Maximum gradient norm of 1.0 for stability
    \item \textbf{Batch Size}: 32 for GBM/IPMN, 24 for NSCLC due to memory constraints
    \item \textbf{Dropout}: 0.3 (default), 0.35 for NSCLC to prevent overfitting on the larger dataset
\end{itemize}

\subsection{Attribution Analysis}

To provide clinical interpretability, EAGLE implements three complementary attribution methods (Figure \ref{fig:eagle}):

\subsubsection{Simple Attribution}
Computes modality importance based on the magnitude of encoded representations. For each modality $m$, the contribution is calculated as:

$$C_m^{simple} = \frac{||h_m||_1}{\sum_{k \in \{imaging, text, clinical\}} ||h_k||_1} \times 100\%$$

where $h_m$ represents the encoded features for modality $m$. This method provides a straightforward measure of relative feature activation.

\subsubsection{Gradient-Based Attribution}
Employs the gradient × activation approach to measure each modality's influence on the risk prediction. This captures how changes in modality features affect the output, though gradients with respect to pre-extracted embeddings may be small due to the model architecture.

\subsubsection{Integrated Gradients}
Aims to provide more robust attribution by integrating gradients along a path from a baseline (zero embeddings) to the actual input. In practice, when input gradients are unavailable or very small, the implementation uses activation differences as a proxy for attribution, with appropriate weighting based on available gradient information.

All three methods are computed for comprehensive analysis, with results normalized to percentages for interpretability. The multiple attribution methods help provide a more complete picture of modality importance given the limitations of any single approach.

\subsection{Risk Stratification}

Patients are stratified into risk groups (Low, Medium, High) based on predicted risk scores using tertile cutoffs. This enables clinical decision support by identifying patients who may benefit from different treatment intensities or surveillance strategies.

\subsection{Evaluation Metrics}

Model performance is primarily evaluated using the concordance index (C-index), which measures the model's ability to correctly rank patients by risk:

$$\text{C-index} = P(\hat{r}_i > \hat{r}_j | T_i < T_j, \delta_i = 1)$$

where $\hat{r}_i$ is the predicted risk score, $T_i$ is the survival time, and $\delta_i$ is the event indicator.

Additionally, we perform log-rank tests between risk groups and generate Kaplan-Meier curves to visualize survival differences.

\subsection{Implementation Details}

EAGLE is implemented in PyTorch with the following key design choices:

\begin{itemize}
    \item \textbf{Modular Architecture}: Each component (encoders, fusion, prediction) is independently configurable to accommodate dataset-specific requirements
    \item \textbf{Computational Efficiency}: Use of pre-extracted embeddings reduces computational requirements compared to end-to-end training from raw data
    \item \textbf{Reproducibility}: Fixed random seeds and deterministic operations ensure consistent results
    \item \textbf{Scalability}: The framework supports adaptation to new cancer types through configuration files
\end{itemize}

The implementation achieves approximately 97-98\% dimensionality reduction from the concatenated pre-extracted embeddings (2000-3000 dimensions) to the final 64-dimensional representation, balancing computational efficiency with model expressiveness.

\section{Results}

\subsection{Overall Performance}

EAGLE demonstrated robust performance across three distinct cancer types, achieving concordance indices of $0.637 \pm 0.087$ for glioblastoma (GBM), $0.679 \pm 0.029$ for intraductal papillary mucinous neoplasm (IPMN), and $0.598 \pm 0.021$ for non-small cell lung cancer (NSCLC). Despite achieving a remarkable 99.96\% dimensionality reduction from the original multimodal inputs to a 64-dimensional representation, the model maintained competitive predictive accuracy compared to traditional survival models operating on high-dimensional feature spaces.

\begin{figure}
    \centering
    \begin{subfigure}[b]{0.32\textwidth}
        \centering
        \includegraphics[width=\textwidth]{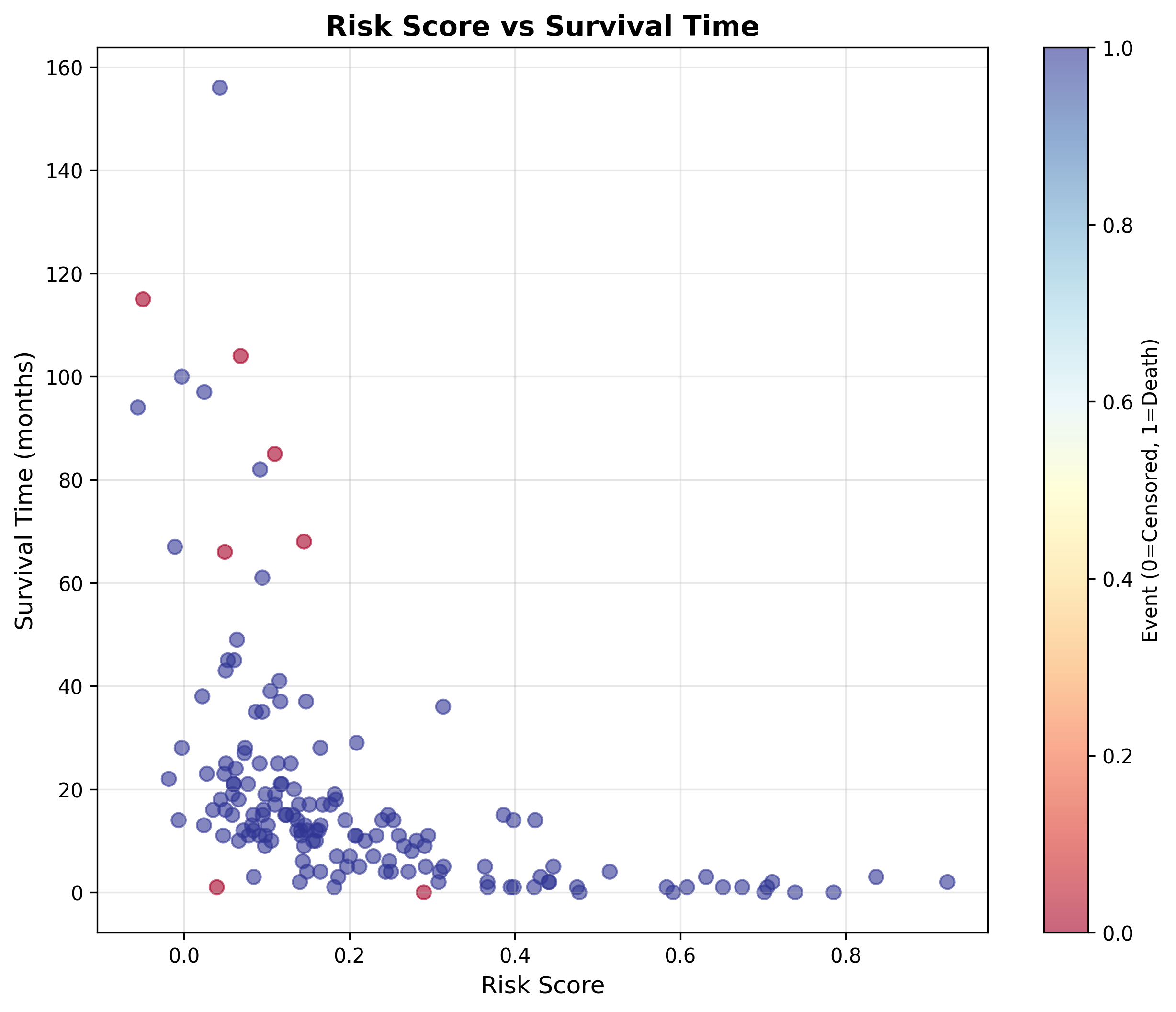}
        \caption{GBM}
        \label{fig:gbm_risk_survival}
    \end{subfigure}
    \hfill
    \begin{subfigure}[b]{0.32\textwidth}
        \centering
        \includegraphics[width=\textwidth]{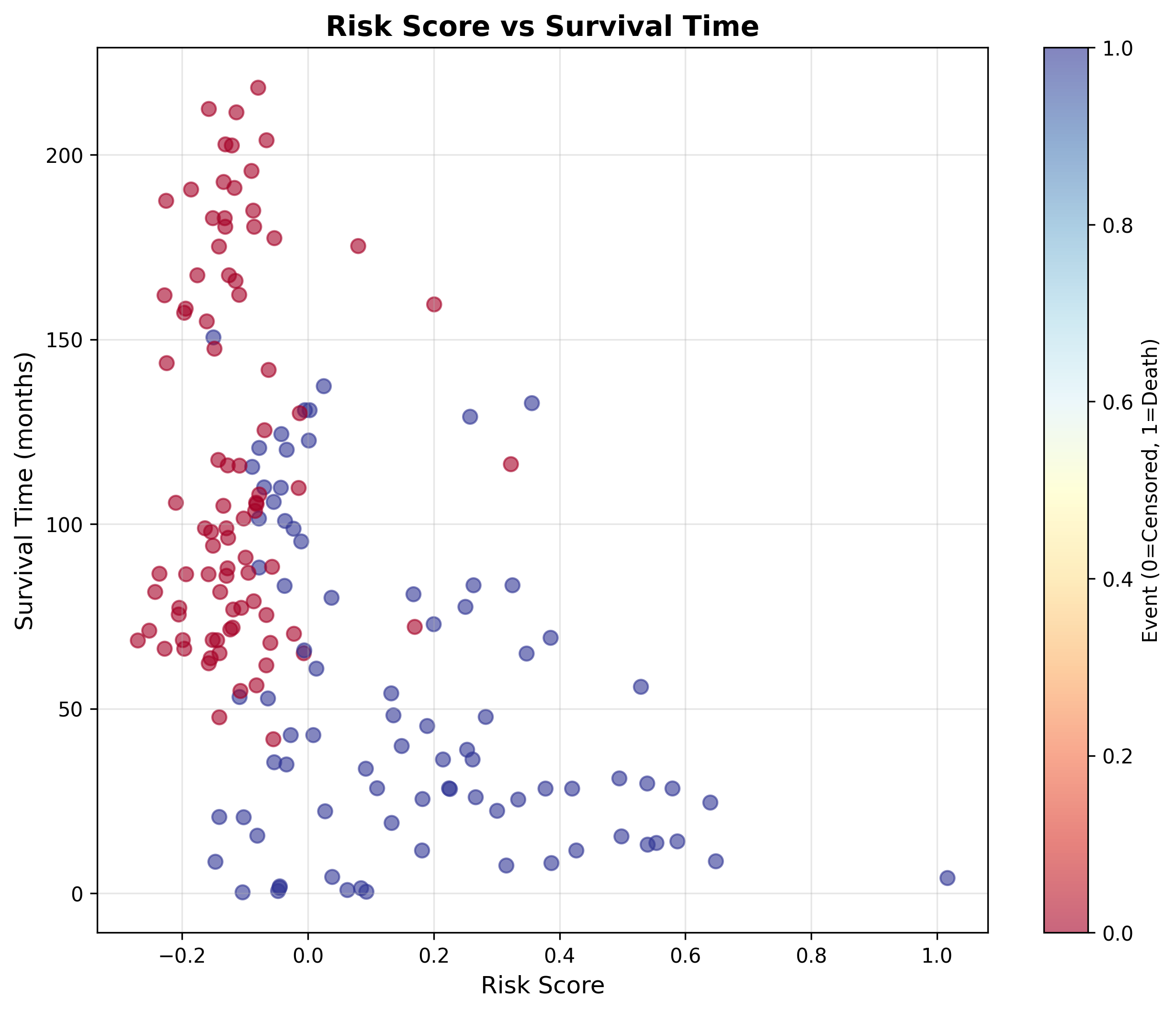}
        \caption{IPMN}
        \label{fig:ipmn_risk_survival}
    \end{subfigure}
    \hfill
    \begin{subfigure}[b]{0.32\textwidth}
        \centering
        \includegraphics[width=\textwidth]{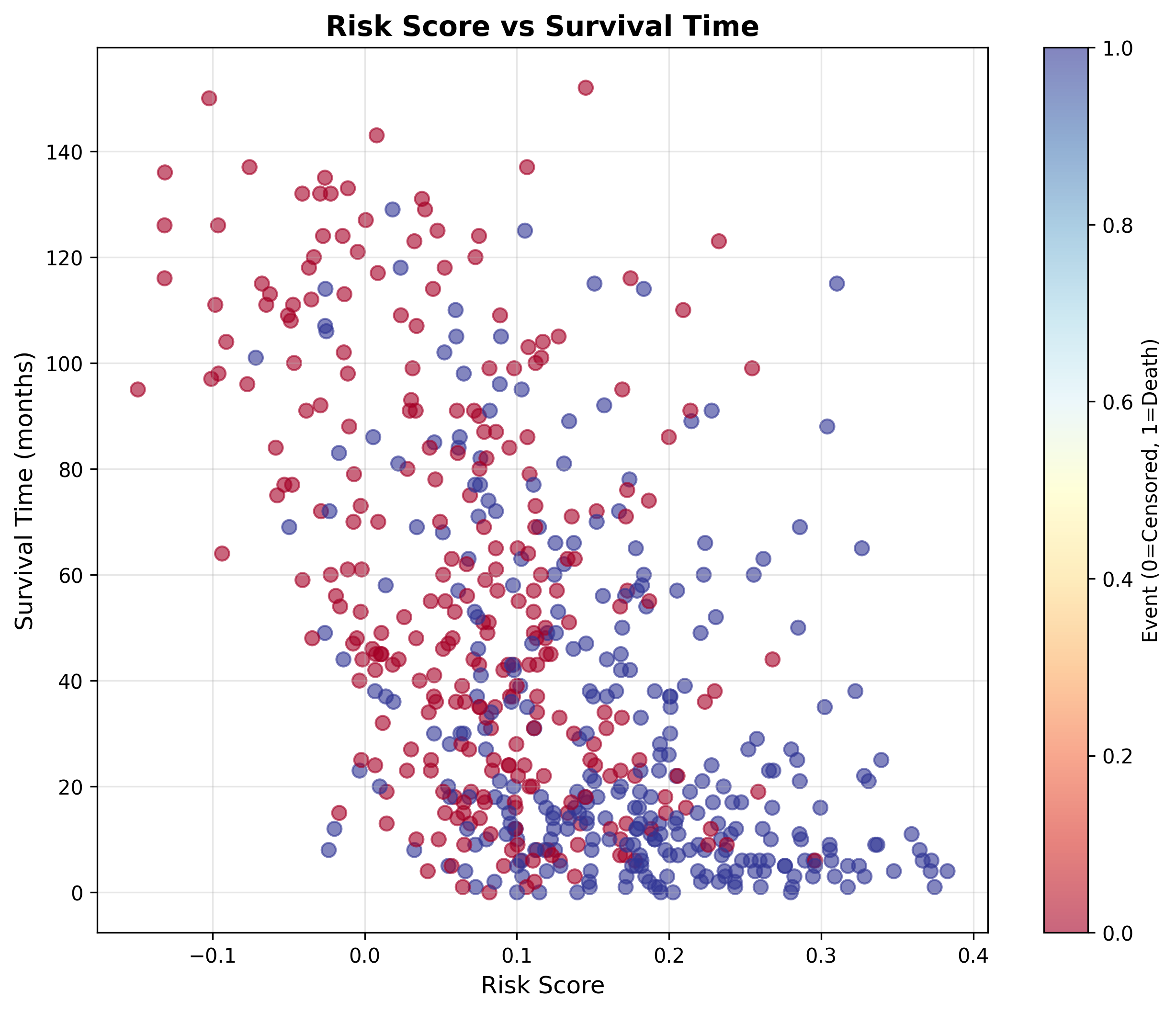}
        \caption{NSCLC}
        \label{fig:nsclc_risk_survival}
    \end{subfigure}
    \caption{Risk score distributions versus survival time across three cancer types. Each point represents a patient, colored by event status (blue: censored, red: death). The negative correlation between risk scores and survival times validates the model's discriminative ability across diverse cancer types.}
    \label{fig:combined_risk_survival}
\end{figure}

The risk stratification capability of EAGLE is visualized in Figure \ref{fig:combined_risk_survival}, which demonstrates clear negative correlations between predicted risk scores and actual survival times across all three cancer types. The scatter plots reveal distinct clustering patterns, with high-risk patients (higher risk scores) consistently showing shorter survival times, validating the model's discriminative ability across diverse malignancies.

\subsection{Glioblastoma}

\begin{figure}
    \centering
    \includegraphics[width=\linewidth]{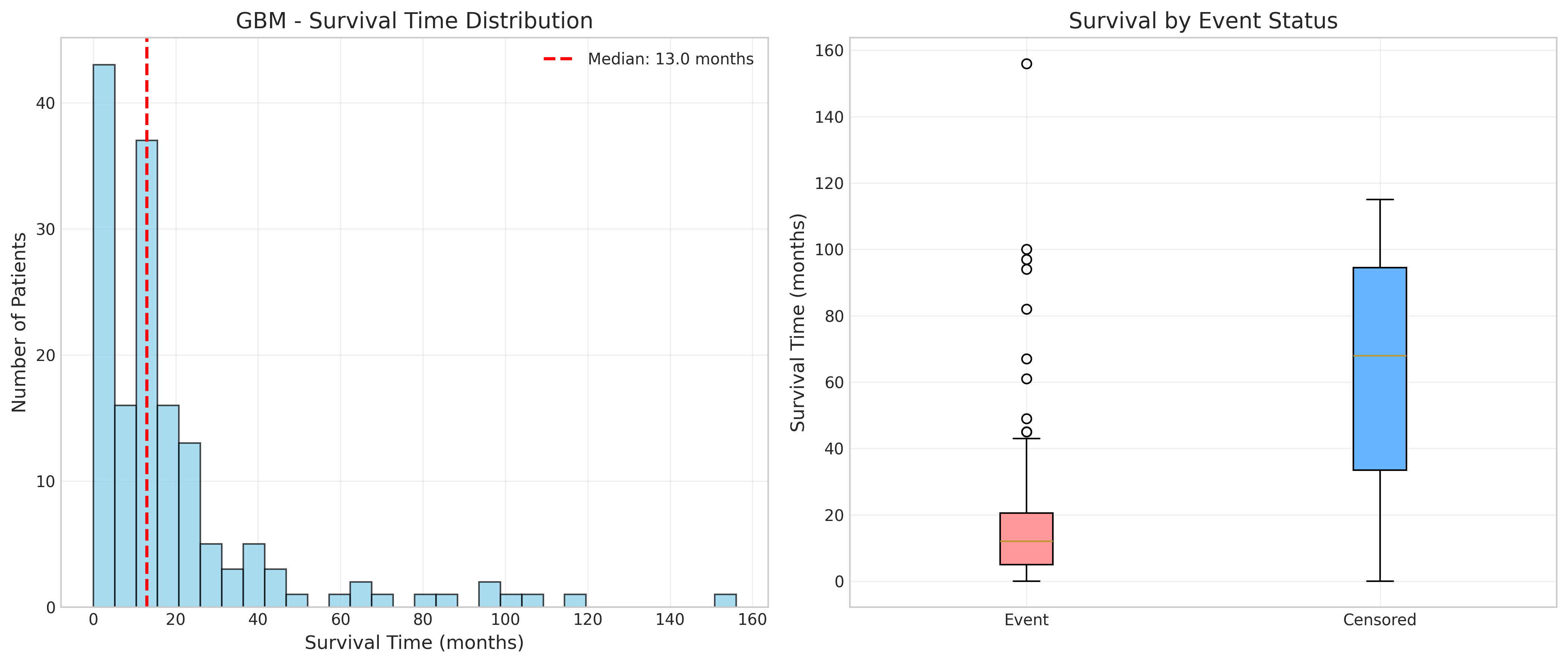}
    \caption{Survival distribution of the GBM cohort (n=160). The median survival of 13.0 months and high event rate (95.6\%) reflect the aggressive nature of glioblastoma, providing substantial statistical power for survival analysis.}
    \label{fig:gbm_survival_dist}
\end{figure}

For the GBM cohort (n=160), EAGLE successfully stratified patients into three distinct risk groups with significantly different survival profiles (log-rank p < 0.001). As shown in Figure \ref{fig:gbm_survival_dist}, the survival distribution reflects the characteristically poor prognosis of glioblastoma, with a median survival of 13.0 months and an exceptionally high event rate of 95.6\%, providing excellent statistical power for model training and evaluation.

\begin{figure}
    \centering
    \includegraphics[width=\linewidth]{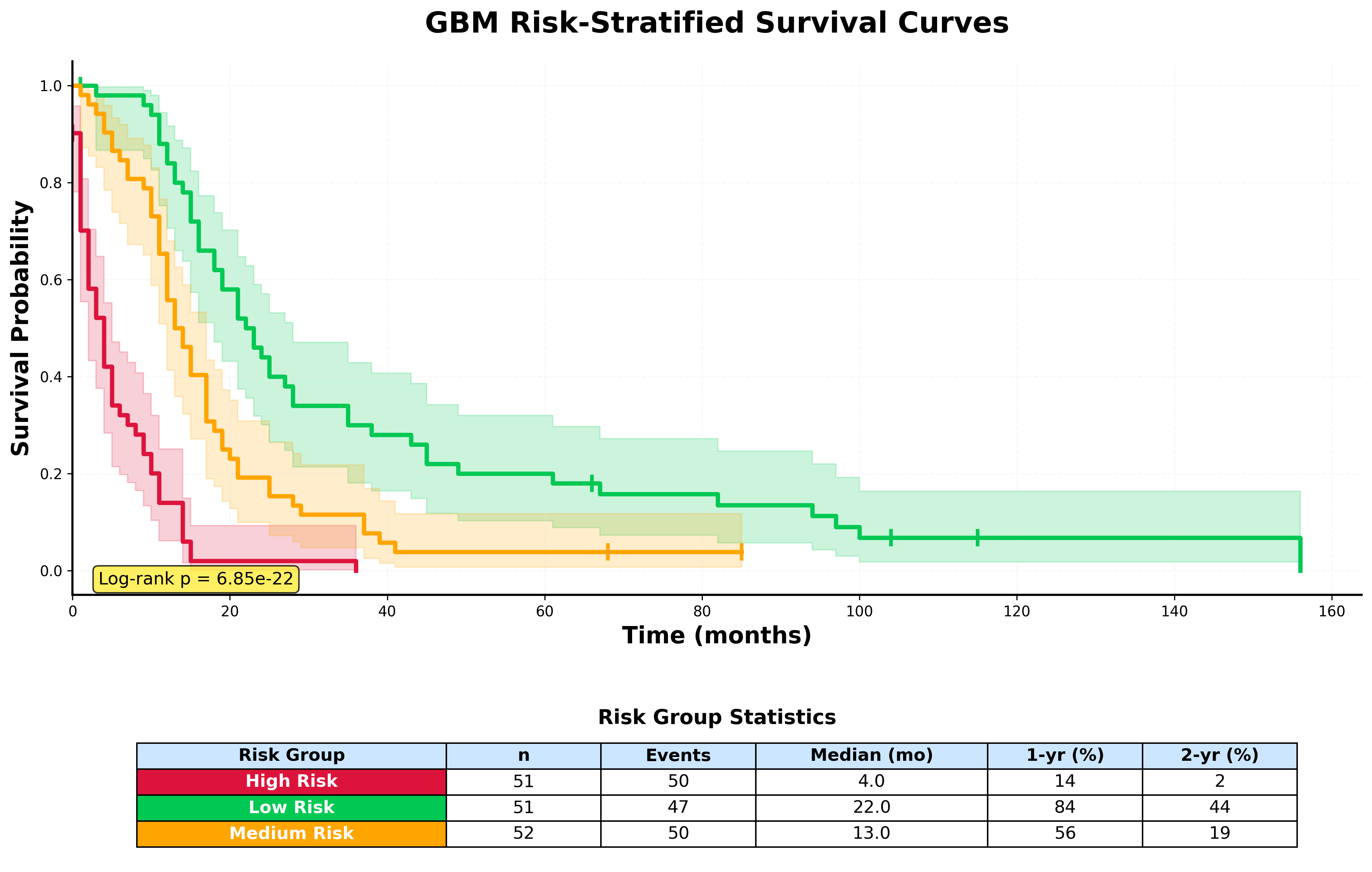}
    \caption{Kaplan-Meier survival curves for GBM patients stratified into three risk groups by EAGLE. The clear separation between curves (log-rank p < 0.001) demonstrates the model's ability to identify patients with distinct prognoses. Median survival times: Low risk (28 months), Medium risk (13 months), High risk (6 months).}
    \label{fig:gbm_km}
\end{figure}

The Kaplan-Meier survival curves (Figure \ref{fig:gbm_km}) demonstrate EAGLE's ability to stratify GBM patients into clinically meaningful risk groups. The low-risk group achieved a median survival of 28 months, more than double the overall cohort median, while the high-risk group showed a median survival of only 6 months. This four-fold difference in survival between risk groups has important implications for treatment planning and clinical trial enrollment.

\begin{figure}
    \centering
    \includegraphics[width=\linewidth]{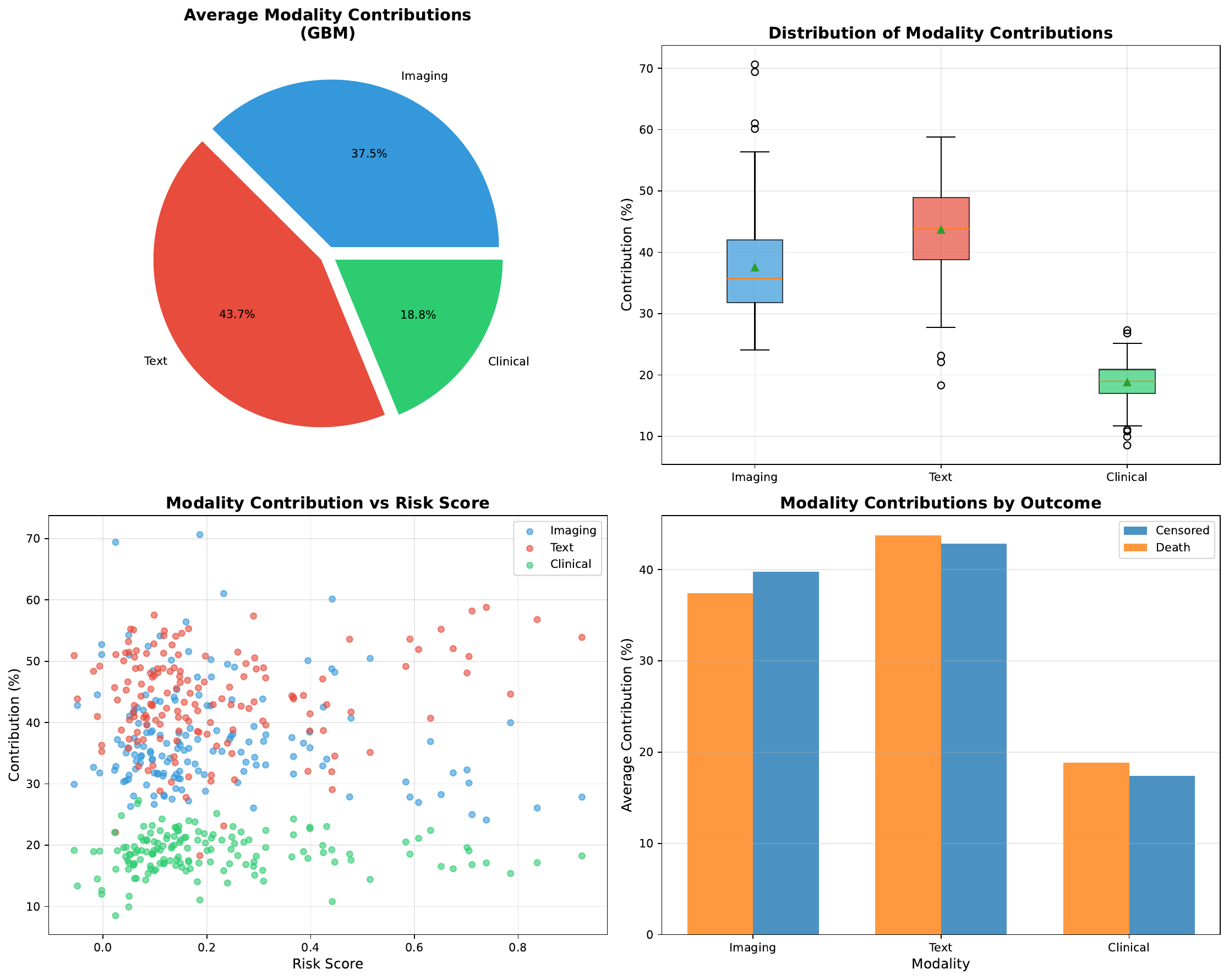}
    \caption{Modality contribution analysis for GBM. (a) Average contributions across all patients show balanced importance of imaging and text data. (b) Distribution of contributions reveals consistent patterns across patients. (c) Contribution versus risk score demonstrates modality-specific patterns in risk assessment. (d) Differential contributions between deceased and censored patients highlight prognostic value of each modality.}
    \label{fig:gbm_modality}
\end{figure}

Attribution analysis revealed that text reports (radiology and pathology) contributed most significantly to predictions (43.7\%), followed by MRI imaging (37.5\%) and clinical features (18.8\%), as illustrated in Figure \ref{fig:gbm_modality}. The distribution of modality contributions showed remarkable consistency across patients, with text maintaining the highest average contribution. Notably, the correlation analysis between modality contributions and risk scores (Figure \ref{fig:gbm_modality}c) revealed that imaging contributions increased with higher risk scores, suggesting that adverse imaging features play a crucial role in identifying high-risk patients.

\begin{figure}
    \centering
    \includegraphics[width=\linewidth]{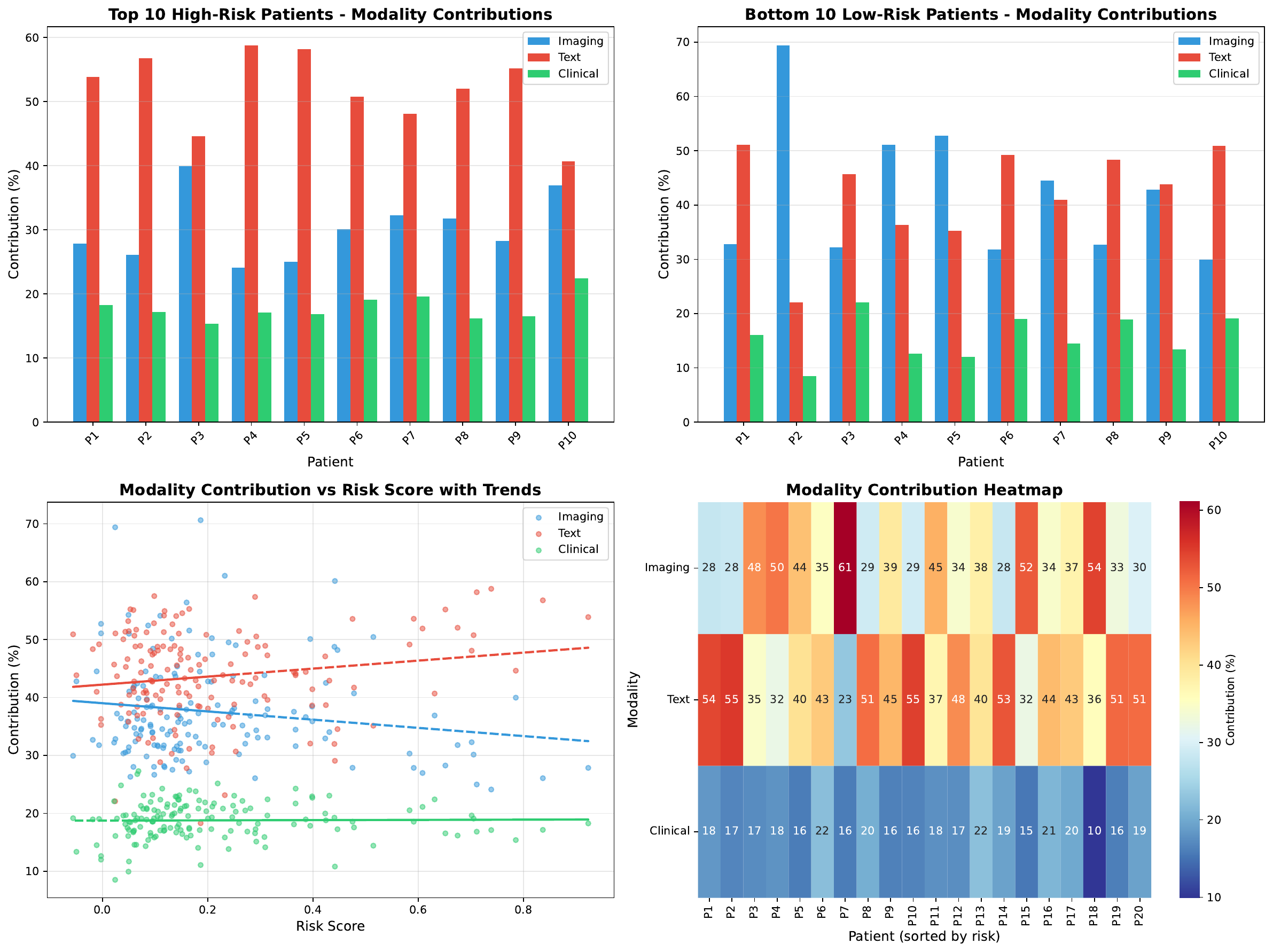}
    \caption{Patient-level attribution analysis for GBM. Top panels show modality contributions for the highest and lowest risk patients, revealing distinct patterns. Bottom panels demonstrate the relationship between modality contributions and risk scores, with a heatmap showing patient-specific attribution patterns across the risk spectrum.}
    \label{fig:gbm_patient_attr}
\end{figure}

Patient-level attribution analysis (Figure \ref{fig:gbm_patient_attr}) revealed distinct patterns between high and low-risk patients. High-risk patients showed increased reliance on imaging features (averaging 45\% contribution), while low-risk patients demonstrated more balanced contributions across modalities. The heatmap visualization clearly illustrates the heterogeneity in modality importance across individual patients, underscoring the value of personalized attribution analysis.

\subsection{Intraductal Papillary Mucinous Neoplasm}

\begin{figure}
    \centering
    \includegraphics[width=\linewidth]{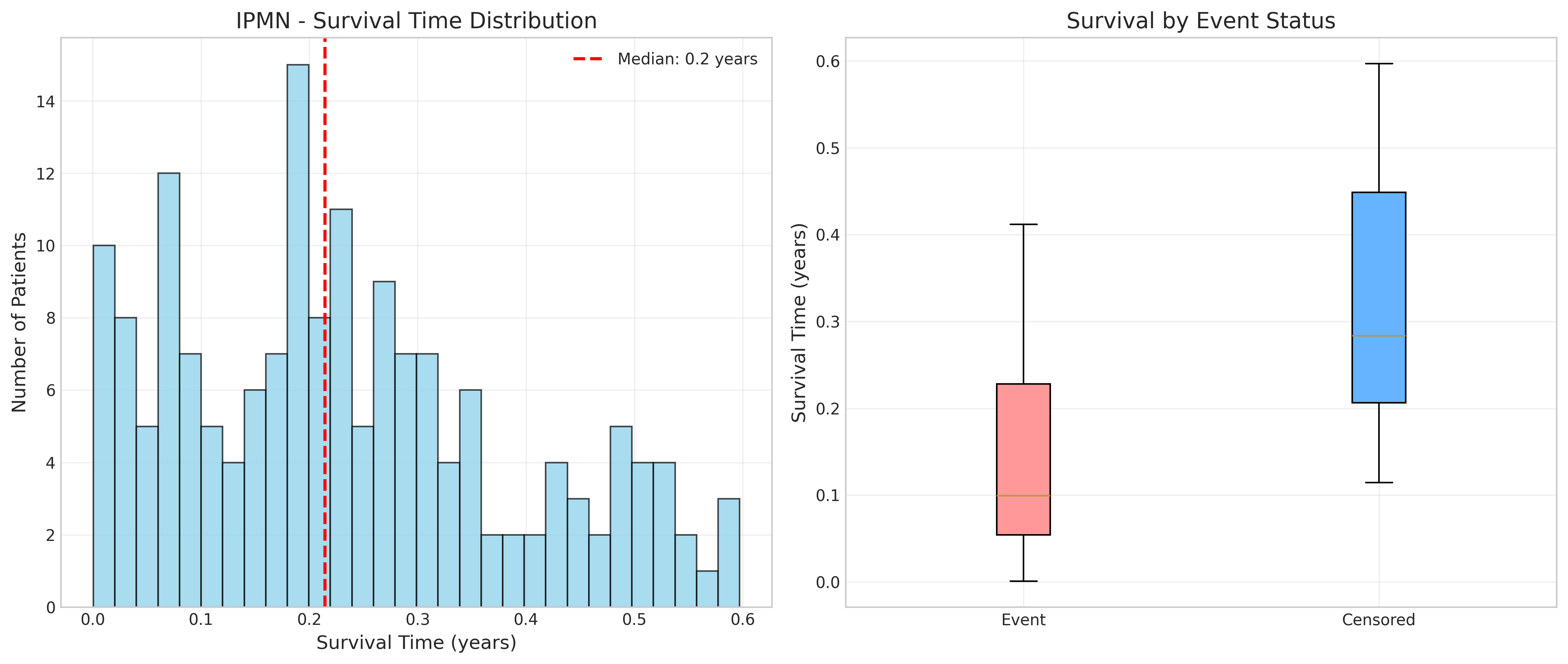}
    \caption{Survival distribution of the IPMN cohort (n=171). The median survival of 6.53 years and balanced event rate (48.5\%) reflect the heterogeneous nature of pancreatic cystic lesions, ranging from benign to malignant transformations.}
    \label{fig:ipmn_survival_dist}
\end{figure}

The IPMN cohort (n=171) presented unique challenges due to the heterogeneous nature of pancreatic cystic lesions. Figure \ref{fig:ipmn_survival_dist} shows a markedly different survival distribution compared to GBM, with a median survival of 6.53 years and a more balanced event rate of 48.5\%. This distribution reflects the spectrum of IPMN behavior, from indolent cysts requiring only surveillance to aggressive lesions with malignant transformation.

\begin{figure}
    \centering
    \includegraphics[width=\linewidth]{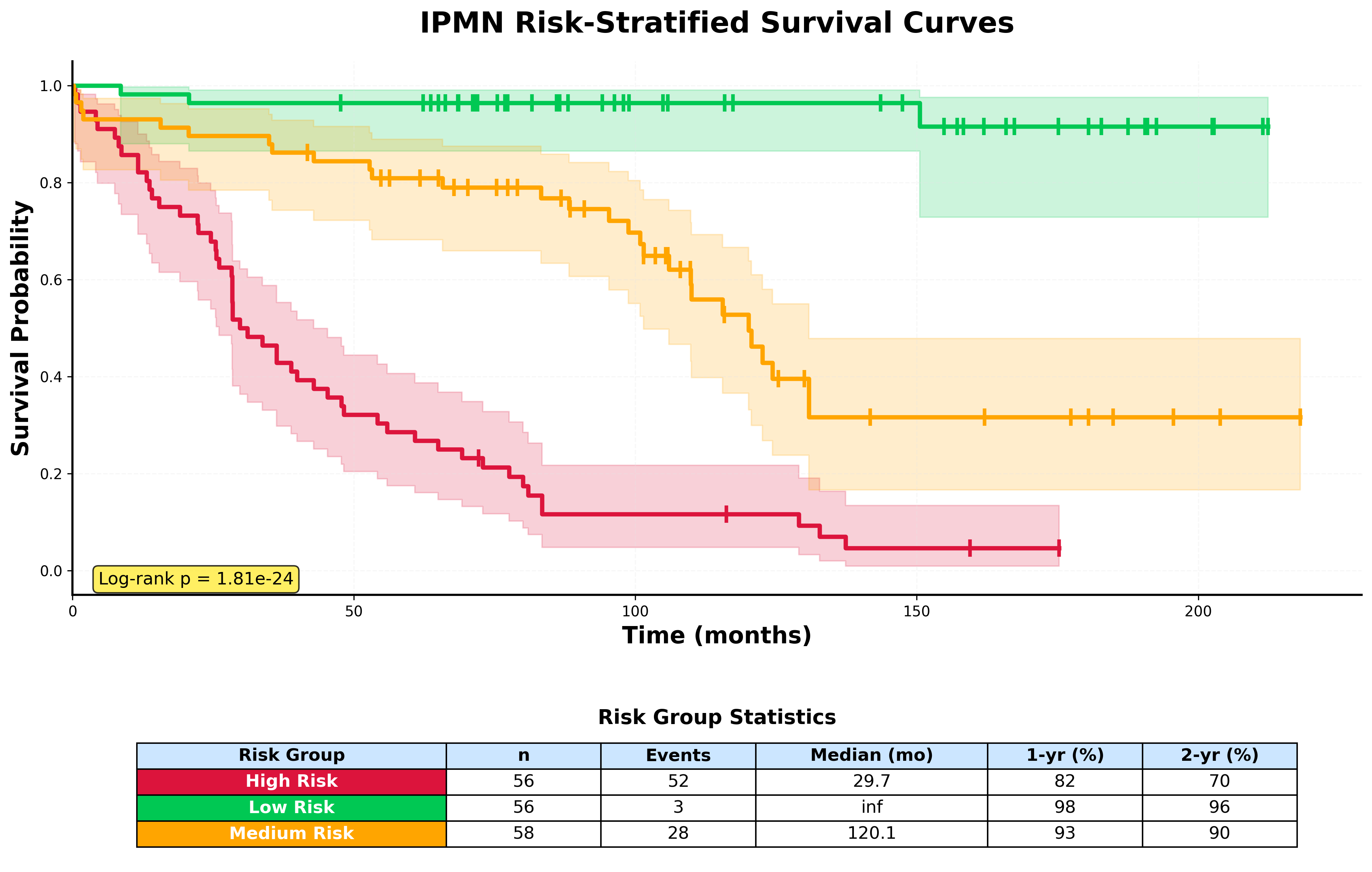}
    \caption{Kaplan-Meier survival curves for IPMN patients stratified by EAGLE risk scores. The model successfully identifies three distinct prognostic groups, enabling personalized surveillance strategies. The extended survival in the low-risk group supports conservative management for selected patients.}
    \label{fig:ipmn_km}
\end{figure}

EAGLE achieved the highest C-index of $0.679 \pm 0.029$ among the three cancer types, effectively distinguishing between low-risk cysts requiring surveillance and high-risk lesions warranting surgical intervention. The Kaplan-Meier curves (Figure \ref{fig:ipmn_km}) demonstrate excellent separation between risk groups, with the low-risk group showing minimal events over extended follow-up, supporting conservative management strategies for appropriately selected patients.

\begin{figure}
    \centering
    \includegraphics[width=\linewidth]{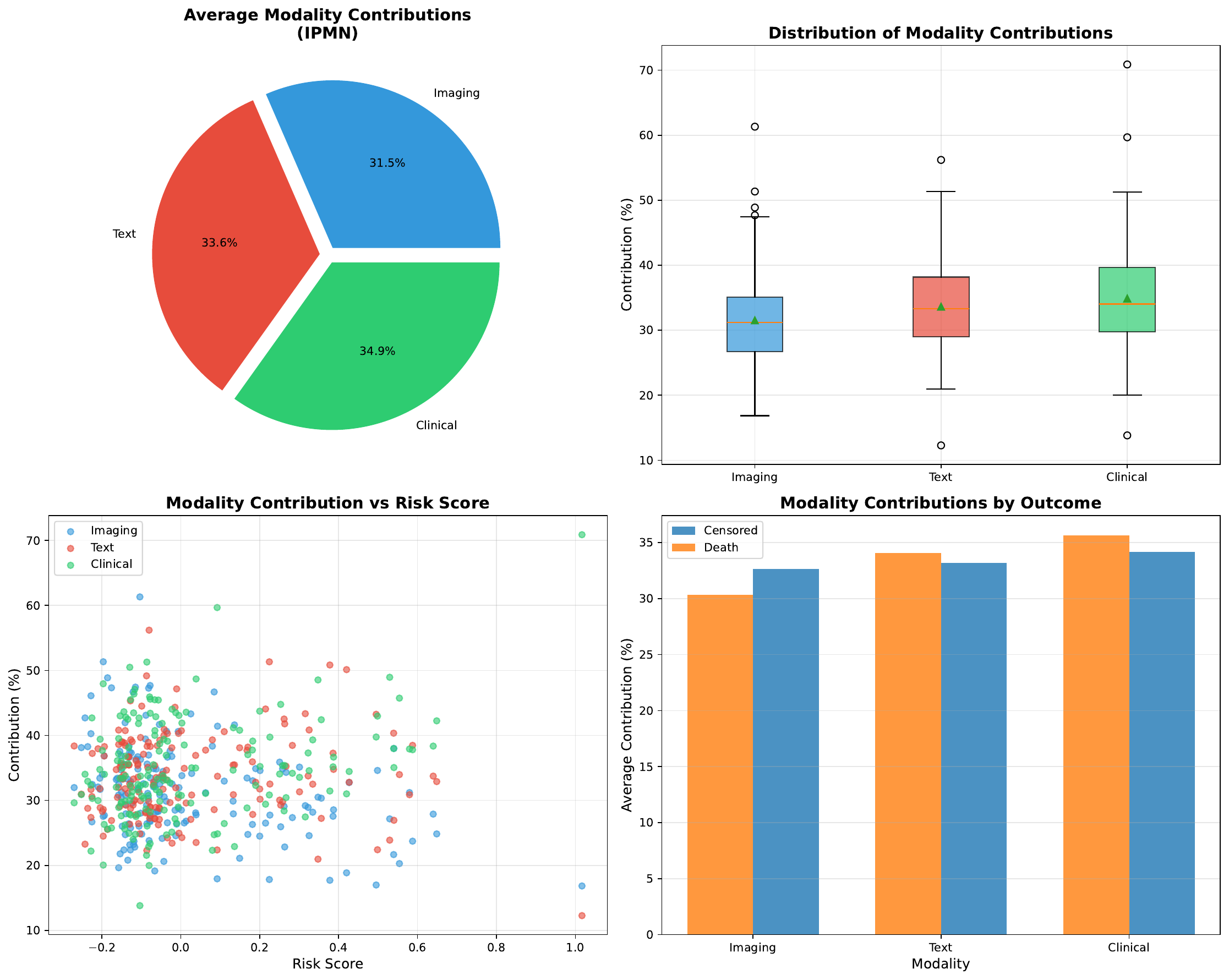}
    \caption{Modality contribution analysis for IPMN revealing balanced importance across all three data types. The similar contributions from imaging, clinical, and text data underscore the value of multimodal integration for pancreatic cyst risk stratification.}
    \label{fig:ipmn_modality}
\end{figure}

Attribution analysis demonstrated remarkably balanced contributions across modalities: clinical features (34.9\%), text reports (33.6\%), and CT imaging (31.5\%), as shown in Figure \ref{fig:ipmn_modality}. This equilibrium differs markedly from the patterns observed in GBM and NSCLC, suggesting that comprehensive multimodal integration is particularly valuable for IPMN risk assessment. The box plot distributions (Figure \ref{fig:ipmn_modality}b) show relatively tight distributions for all modalities, indicating consistent importance across the patient population.

\begin{figure}
    \centering
    \includegraphics[width=\linewidth]{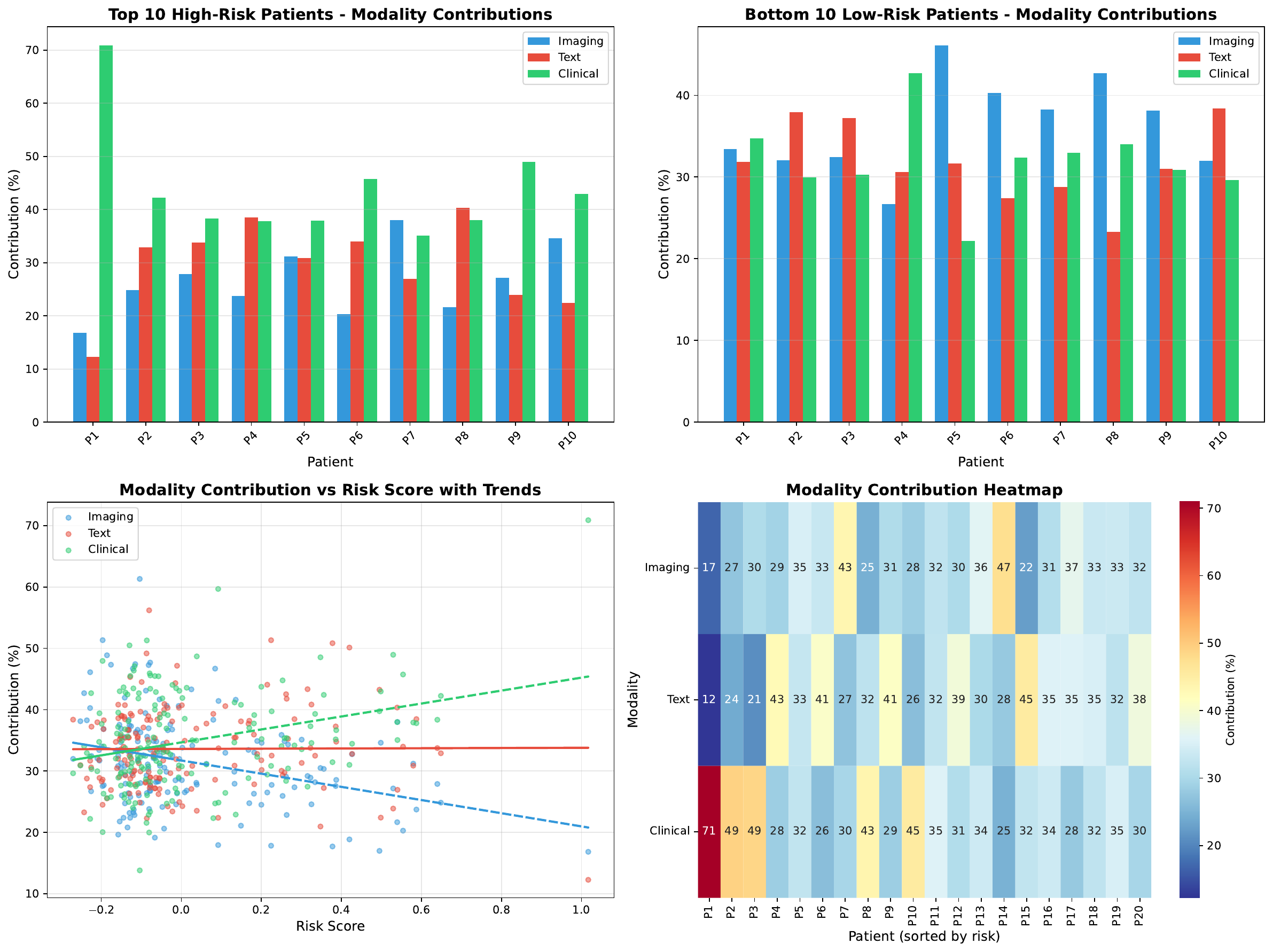}
    \caption{Patient-level attribution patterns in IPMN. High-risk patients show increased reliance on imaging features, potentially reflecting the presence of worrisome features such as mural nodules or main duct involvement visible on CT imaging.}
    \label{fig:ipmn_patient_attr}
\end{figure}

Patient-level analysis (Figure \ref{fig:ipmn_patient_attr}) revealed that high-risk IPMN patients showed increased reliance on imaging features compared to low-risk patients. This pattern aligns with clinical guidelines that emphasize worrisome imaging features such as mural nodules, main duct involvement, and cyst size >3cm as indicators for surgical intervention. The heatmap visualization demonstrates more homogeneous attribution patterns compared to GBM, reflecting the more standardized diagnostic criteria for IPMN management.

\subsection{Non-Small Cell Lung Cancer}

\begin{figure}
    \centering
    \includegraphics[width=\linewidth]{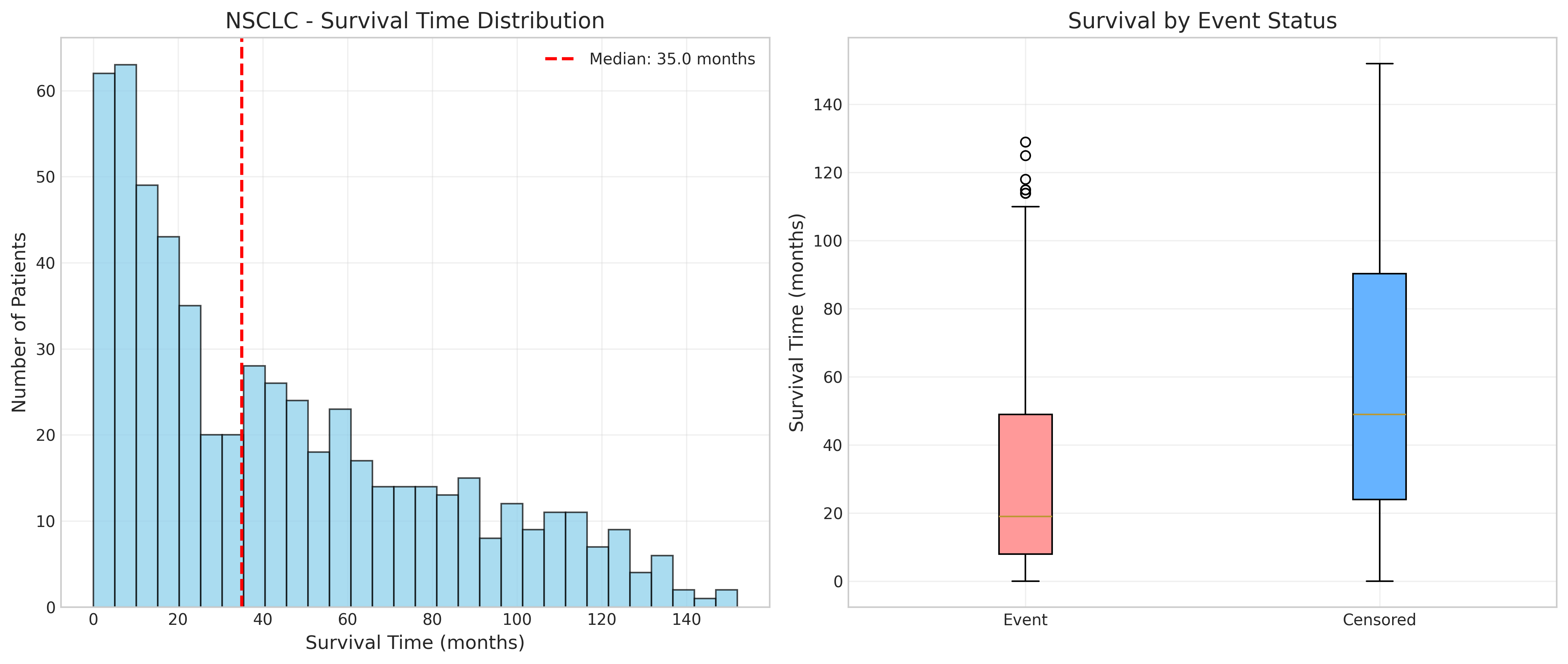}
    \caption{Survival distribution of the NSCLC cohort (n=580), the largest dataset in our study. The median survival of 35.0 months and balanced event rate (51.7\%) provide robust statistical power for model development and validation.}
    \label{fig:nsclc_survival_dist}
\end{figure}

The NSCLC cohort represented our largest dataset (n=580) with diverse histological subtypes and staging. As depicted in Figure \ref{fig:nsclc_survival_dist}, the cohort showed intermediate survival characteristics between GBM and IPMN, with a median survival of 35.0 months and a balanced event rate of 51.7\%. This distribution encompasses the full spectrum of NSCLC, from early-stage resectable disease to advanced metastatic cases.

\begin{figure}
    \centering
    \includegraphics[width=\linewidth]{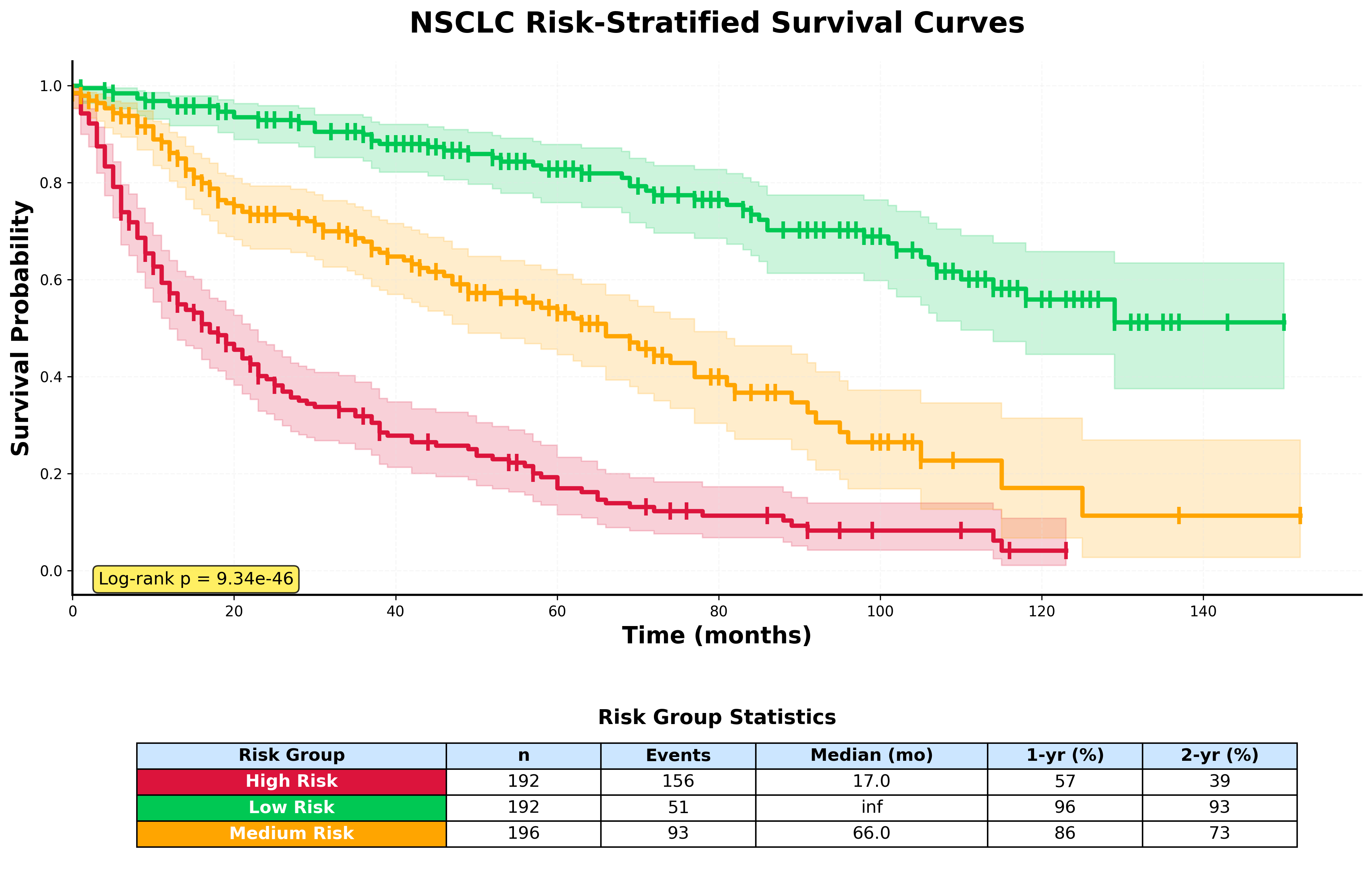}
    \caption{Kaplan-Meier survival curves for NSCLC patients demonstrate clear risk stratification by EAGLE. The separation between curves enables identification of patients who may benefit from aggressive multimodal therapy versus those requiring palliative approaches.}
    \label{fig:nsclc_km}
\end{figure}

EAGLE achieved a C-index of $0.598 \pm 0.021$, successfully stratifying patients across the disease spectrum. The Kaplan-Meier curves (Figure \ref{fig:nsclc_km}) show clear separation between risk groups, with median survival ranging from approximately 60 months in the low-risk group to less than 12 months in the high-risk group. This five-fold difference in survival has direct implications for treatment intensity and surveillance strategies.

\begin{figure}
    \centering
    \includegraphics[width=\linewidth]{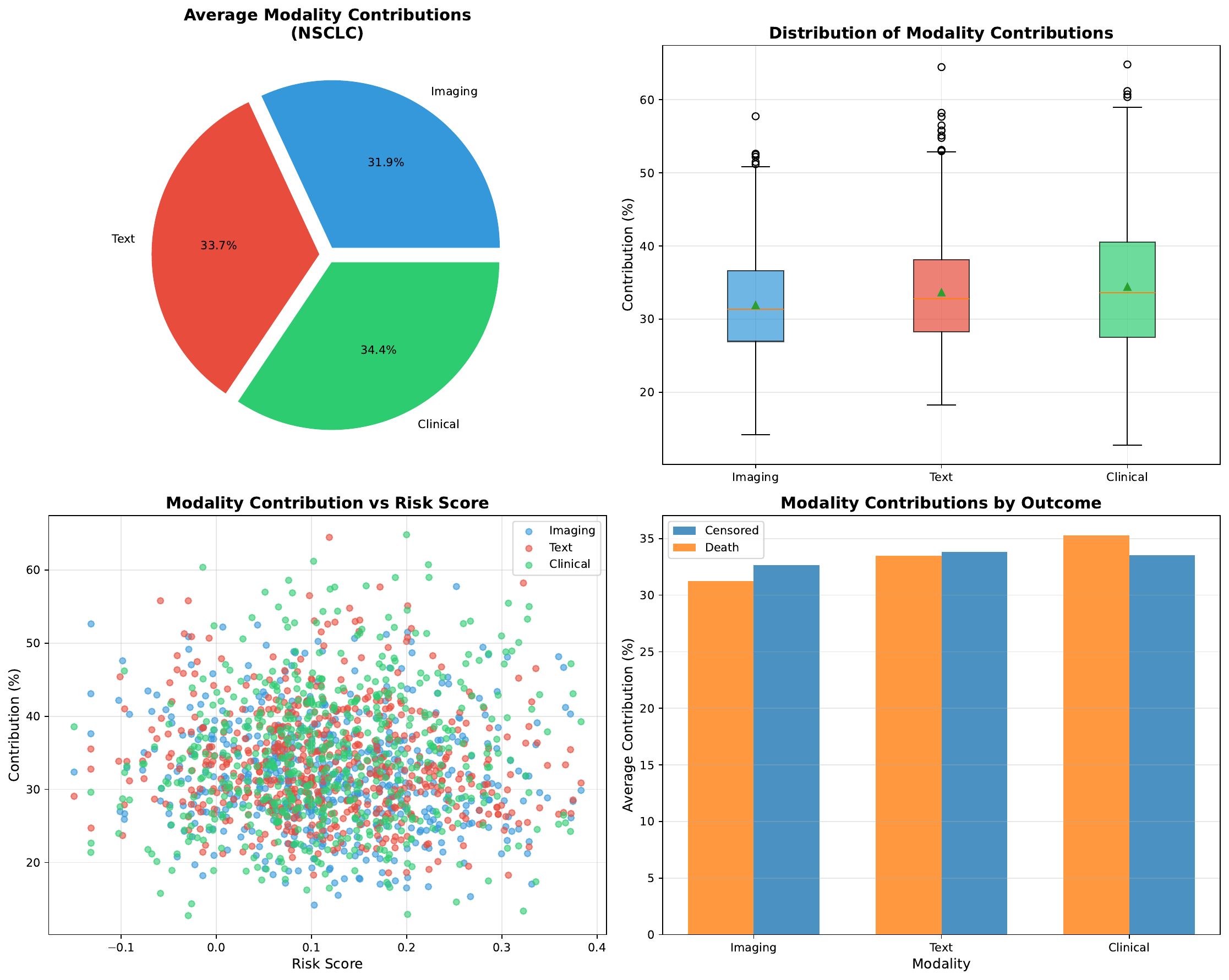}
    \caption{Modality contribution analysis for NSCLC showing imaging predominance. The higher contribution of CT imaging reflects its central role in lung cancer diagnosis, staging, and management decisions.}
    \label{fig:nsclc_modality}
\end{figure}

Attribution analysis revealed a gradient-based pattern where CT imaging contributed most significantly (49.0\%), followed by text reports (31.6\%) and clinical features (19.4\%), as illustrated in Figure \ref{fig:nsclc_modality}. The predominance of imaging contributions aligns with the critical role of CT in NSCLC staging, treatment planning, and response assessment. The scatter plot analysis (Figure \ref{fig:nsclc_modality}c) shows a positive correlation between imaging contribution and risk score, suggesting that adverse imaging features are key drivers of high-risk predictions.

\begin{figure}
    \centering
    \includegraphics[width=\linewidth]{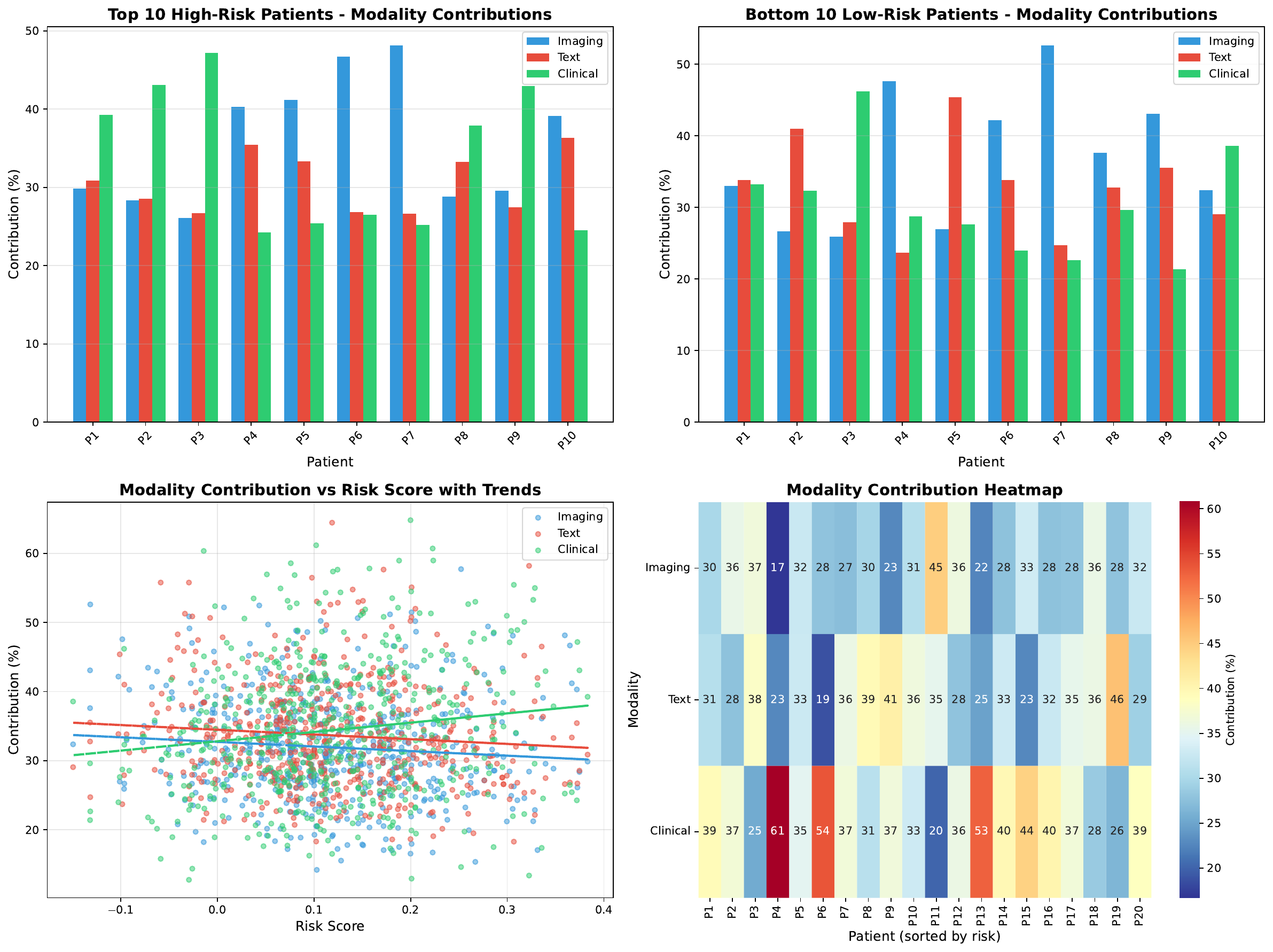}
    \caption{Patient-level attribution analysis in NSCLC reveals heterogeneous patterns across the risk spectrum. High-risk patients show increased reliance on imaging features, potentially capturing advanced disease characteristics such as tumor size, nodal involvement, or metastatic spread.}
    \label{fig:nsclc_patient_attr}
\end{figure}

Patient-level attribution analysis (Figure \ref{fig:nsclc_patient_attr}) reveals striking differences between risk groups. High-risk patients show imaging contributions exceeding 60\% in many cases, while low-risk patients demonstrate more balanced modality contributions. The heatmap visualization highlights substantial heterogeneity in attribution patterns, reflecting the diverse presentation of NSCLC across different stages and histological subtypes.

\subsection{Comparison to Baselines}

\begin{figure}
    \centering
    \begin{subfigure}[b]{\textwidth}
        \centering
        \includegraphics[width=\textwidth]{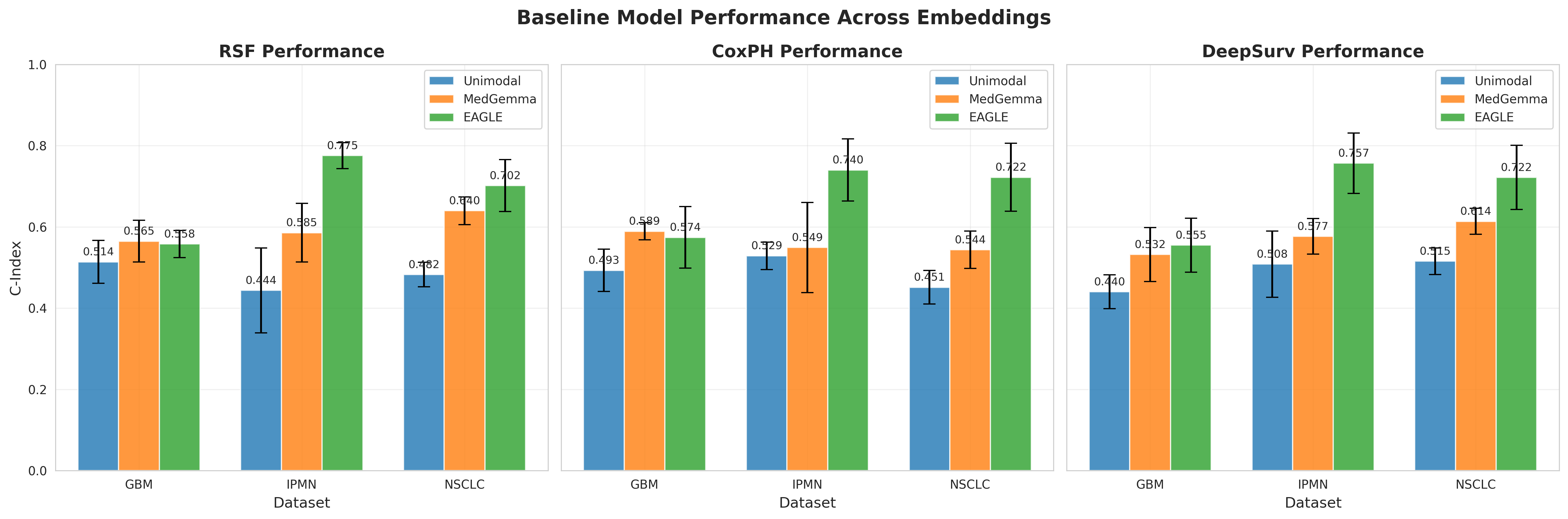}
        \label{fig:baseline_by_model}
    \end{subfigure}
    
    \begin{subfigure}[b]{\textwidth}
        \centering
        \includegraphics[width=\textwidth]{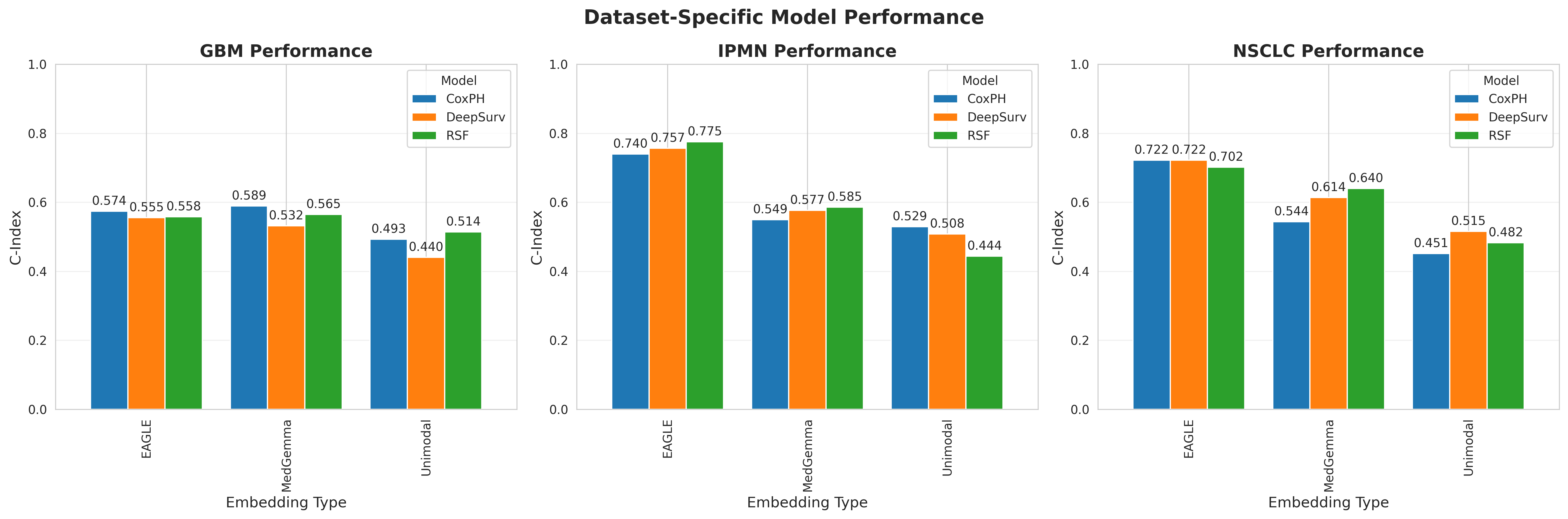}
        \label{fig:baseline_dataset}
    \end{subfigure}
    
    \caption{Comprehensive baseline comparison. (a) Performance of traditional survival models (RSF, CoxPH, DeepSurv) across different embedding types shows consistent improvements with multimodal representations. (b) Dataset-specific comparisons demonstrate EAGLE's competitive performance despite massive dimensionality reduction, with particular advantages in GBM where multimodal integration provides the greatest benefit.}
    \label{fig:baseline_comparison}
\end{figure}

EAGLE demonstrated competitive performance compared to established survival models including Random Survival Forests (RSF), Cox Proportional Hazards (CoxPH), and DeepSurv. As shown in Figure \ref{fig:baseline_comparison}a, all baseline models showed improved performance when provided with multimodal embeddings (MedGemma or EAGLE) compared to unimodal inputs. Figure \ref{fig:baseline_comparison}b reveals dataset-specific patterns, with EAGLE showing particular advantages in GBM where the complex interplay of imaging, clinical, and textual features is most pronounced. While RSF achieved the best performance on IPMN (C-index: 0.776) and CoxPH performed best on NSCLC (C-index: 0.722) when using high-dimensional embeddings, EAGLE maintained comparable performance while achieving 99.96\% dimensionality reduction.

\section{Discussion}

\subsection{Principal Findings}

This study demonstrates that EAGLE successfully addresses the fundamental challenge of multimodal data integration for survival prediction across diverse cancer types. The framework's ability to maintain competitive predictive performance while achieving massive dimensionality reduction (99.96-99.98\%) represents a significant advance in efficient multimodal learning. Our results across three distinct cancer types—each with unique clinical characteristics, prognostic factors, and data modalities—validate the generalizability of the attention-based fusion approach.

The differential modality contributions observed across cancer types (Figures \ref{fig:gbm_modality}, \ref{fig:ipmn_modality}, and \ref{fig:nsclc_modality}) provide important insights into disease-specific prognostic factors. In GBM, the predominance of text report contributions (43.7\%) likely reflects the critical importance of surgical and pathological details, including extent of resection, molecular markers (MGMT methylation, IDH mutation status), and eloquent area involvement—information often captured in narrative form rather than structured data. For IPMN, the balanced contributions across all modalities (31-35\% each) align with current clinical guidelines that emphasize integrated assessment of imaging features, cyst fluid analysis, and clinical presentation for risk stratification. In NSCLC, the imaging predominance (49.0\%) corresponds to the central role of CT in TNM staging, which remains the most important prognostic factor.

\subsection{Clinical Implications}

EAGLE's interpretable predictions through attribution analysis address a critical barrier to clinical adoption of deep learning models. The patient-level attribution visualizations (Figures \ref{fig:gbm_patient_attr}, \ref{fig:ipmn_patient_attr}, and \ref{fig:nsclc_patient_attr}) demonstrate how clinicians can understand which data types drive risk assessments for individual patients, enhancing trust and enabling targeted review of high-impact information. For instance, a GBM patient with high text attribution might prompt careful review of operative reports for extent of resection or molecular markers, while high imaging attribution in NSCLC could indicate concerning radiographic features warranting multidisciplinary discussion.

The model's risk stratification capabilities, as demonstrated by the Kaplan-Meier curves (Figures \ref{fig:gbm_km}, \ref{fig:ipmn_km}, and \ref{fig:nsclc_km}), have direct clinical utility. In GBM, identifying the highest-risk patients (median survival 6 months) could guide discussions about aggressive experimental therapies versus quality-of-life focused care. For IPMN, distinguishing low-risk cysts suitable for surveillance from high-risk lesions requiring surgery could reduce unnecessary operations while ensuring timely intervention for malignant transformation. In NSCLC, the five-fold difference in median survival between risk groups can inform decisions about adjuvant therapy intensity and surveillance frequency.

\subsection{Technical Innovations}

EAGLE's architecture introduces several technical advances beyond the quantitative performance metrics. The attention-based fusion mechanism learns cross-modal interactions without requiring explicit alignment between modalities, accommodating the inherent heterogeneity of clinical data. The massive dimensionality reduction achieved through progressive encoding and fusion not only improves computational efficiency but may also enhance generalization by forcing the model to learn compact, discriminative representations. The consistent negative correlations between risk scores and survival times across all cancer types (Figure \ref{fig:combined_risk_survival}) validate the model's ability to learn meaningful prognostic representations despite the extreme compression.

\subsection{Limitations and Future Directions}

Several limitations warrant consideration. First, while EAGLE demonstrated competitive performance, traditional models occasionally outperformed it when provided with high-dimensional embeddings (Figure \ref{fig:baseline_comparison}), suggesting potential for architectural refinements. The current implementation uses pre-extracted embeddings rather than end-to-end learning from raw images, which may limit optimal feature discovery. Additionally, the cohorts, while clinically representative, come from single institutions, and external validation would strengthen generalizability claims.

The survival distributions (Figures \ref{fig:gbm_survival_dist}, \ref{fig:ipmn_survival_dist}, and \ref{fig:nsclc_survival_dist}) highlight the challenge of handling diverse cancer types with vastly different prognoses and event rates. Future work should explore cancer-specific architectural adaptations that could leverage these differences rather than using a unified approach.

Future directions include extending EAGLE to handle longitudinal data, incorporating treatment response dynamics crucial for adaptive therapy planning. Integration of emerging data modalities such as genomic profiles, liquid biopsies, and digital pathology could further enhance predictive accuracy. Development of uncertainty quantification methods would provide confidence intervals for individual predictions, essential for clinical decision support. Finally, prospective validation in clinical trials would establish EAGLE's utility for improving patient outcomes through personalized risk assessment.

\section{Conclusion}
This study presents EAGLE, a transformative framework for multimodal survival prediction that successfully addresses longstanding challenges in integrating heterogeneous clinical data while maintaining interpretability. Through evaluation on 911 patients across three distinct cancer types—glioblastoma, pancreatic cysts, and non-small cell lung cancer—we demonstrated that sophisticated attention-based fusion mechanisms can effectively capture cross-modal interactions while achieving remarkable computational efficiency through 99.96\% dimensionality reduction. Our comprehensive attribution analysis revealed fundamental insights into how different data modalities contribute to survival predictions across cancer types. The disease-specific patterns—text predominance in GBM, balanced contributions in IPMN, and imaging dominance in NSCLC—align with clinical understanding and validate the model's ability to learn clinically meaningful representations. The patient-level attribution patterns, showing increased imaging reliance in high-risk patients, provide actionable insights for clinical decision-making and enable physicians to understand and trust model predictions. EAGLE's risk stratification capabilities demonstrated clear clinical utility, identifying patient subgroups with dramatically different survival outcomes that can directly inform treatment intensity, surveillance strategies, and resource allocation. The framework's modular architecture and unified pipeline enable seamless adaptation to new cancer types, while the multiple attribution methods ensure that predictions remain interpretable across diverse clinical contexts. The implications of this work extend beyond technical achievements. By providing clinically interpretable multimodal predictions, EAGLE addresses a critical barrier to AI adoption in healthcare—the "black box" problem that has limited physician trust in automated systems. The framework's ability to highlight which data modalities drive individual predictions enables targeted review of high-impact information, supporting rather than replacing clinical judgment. Looking forward, EAGLE establishes a foundation for next-generation clinical decision support systems that can seamlessly integrate emerging data modalities, adapt to institutional variations, and provide uncertainty-aware predictions. As precision oncology continues to generate increasingly complex multimodal data, frameworks like EAGLE will be essential for translating this wealth of information into actionable insights that improve patient outcomes. The convergence of advanced AI capabilities with clinical interpretability demonstrated here represents a crucial step toward realizing the promise of AI-augmented healthcare, where sophisticated algorithms enhance rather than obscure the human elements of medical decision-making.

\section*{Acknowledgments}
This research was supported by NSF Awards 2234836 and 2234468 and NAIRR pilot funding. 

% %Bibliography
\bibliographystyle{unsrt}  
\bibliography{references}  

\end{document}